\documentclass[10pt, conference, compsocconf]{IEEEtran}

\usepackage{mathptmx} 
\usepackage{algorithm}
\usepackage{algpseudocode}

\usepackage[hyphens]{url}
\usepackage[sort,nocompress]{cite}
\usepackage[final]{microtype}
\usepackage[keeplastbox]{flushend}
\usepackage{balance}
\usepackage{booktabs}
\usepackage{todonotes}
\usepackage{epsfig}
\usepackage{graphicx}
\usepackage{amsmath}
\usepackage{amssymb}
\usepackage{lipsum}
\usepackage{subfig}
\usepackage{color}
\usepackage{url}

\usepackage[normalem]{ulem}
\usepackage{bigstrut}
\usepackage{multirow}
\usepackage{siunitx}
\usepackage{ulem}
\usepackage{listings}
\usepackage{eso-pic}
\usepackage{tikz}
\usepackage{xspace}
\usepackage{fancyhdr}
\usepackage{txfonts}
\usepackage{mathtools}
\usepackage{soul}
\usepackage{tipa}
\usepackage{textgreek}

\usepackage[colorlinks  = true,
            linkcolor   = blue,
            citecolor   = red,
            urlcolor    = blue,
            anchorcolor = blue,
            breaklinks,
            bookmarks = false,
            draft=false]{hyperref}   

\title{Mesorasi: Taming Redundancies and Embracing Irregularities for Efficient Deep Point Cloud Analytics}
\title{Mesorasi: Enabling Efficient Point Cloud Analytics via Delayed-Aggregation}
\title{Mesorasi: Architecture Support for Point Cloud Analytics via Delayed-Aggregation}

\makeatletter
\newcommand{\linebreakand}{%
  \end{@IEEEauthorhalign}
  \hfill\mbox{}\par
  \mbox{}\hfill\begin{@IEEEauthorhalign}
}
\makeatother

\author{\IEEEauthorblockN{Yu Feng\textsuperscript{*}}
\IEEEauthorblockA{University of Rochester\\yfeng28@ur.rochester.edu}
\and
\IEEEauthorblockN{Boyuan Tian\textsuperscript{*}}
\IEEEauthorblockA{University of Rochester\\btian2@ur.rochester.edu}
\and
\IEEEauthorblockN{Tiancheng Xu\textsuperscript{*}}
\IEEEauthorblockA{University of Rochester\\txu17@ur.rochester.edu}
\linebreakand
\IEEEauthorblockN{Paul Whatmough}
\IEEEauthorblockA{Arm Research\\paul.whatmough@arm.com}
\and
\IEEEauthorblockN{Yuhao Zhu}
\IEEEauthorblockA{University of Rochester\\yzhu@rochester.edu}
\linebreakand
\IEEEauthorblockN{\texttt{\url{horizon-lab.org}}}
}

\ifx\figurename\undefined \def\figurename{Figure}\fi
\renewcommand{\figurename}{Fig.}
\renewcommand{\paragraph}[1]{\textbf{#1} }

\newcommand{\Sect}[1]{Sec.~\ref{#1}}
\newcommand{\Fig}[1]{Fig.~\ref{#1}}
\newcommand{\Tbl}[1]{Tbl.~\ref{#1}}
\newcommand{\Equ}[1]{Equ.~\ref{#1}}

\newcommand{\specialcell}[2][c]{\begin{tabular}[#1]{@{}c@{}}#2\end{tabular}}

\newcommand{\proj}{\textsc{Mesorasi}\xspace}

\newcommand{\sys}[1]{\underline{\textsc{#1}}}

\newcommand{\no}[1]{#1}
\renewcommand{\no}[1]{}
\newcommand{\RNum}[1]{\uppercase\expandafter{\romannumeral #1\relax}}

\def\cA{{\mathcal{A}}}
\def\cF{{\mathcal{F}}}
\def\cN{{\mathcal{N}}}

\graphicspath{{figs/}}

\begin{document}
\maketitle
\begingroup\renewcommand\thefootnote{*}
\footnotetext{Equal contribution}
\endgroup


\begin{abstract}


Point cloud analytics is poised to become a key workload on battery-powered embedded and mobile platforms in a wide range of emerging application domains, such as autonomous driving, robotics, and augmented reality, where efficiency is paramount. This paper proposes \proj, an algorithm-architecture co-designed system that simultaneously improves the performance and energy efficiency of point cloud analytics while retaining its accuracy.


Our extensive characterizations of state-of-the-art point cloud algorithms show that, while structurally reminiscent of convolutional neural networks (CNNs),  point cloud algorithms exhibit inherent compute and memory inefficiencies due to the unique characteristics of point cloud data. We propose \textit{delayed-aggregation}, a new algorithmic primitive for building efficient point cloud algorithms. Delayed-aggregation hides the performance bottlenecks and reduces the compute and memory redundancies by exploiting the \textit{approximately distributive} property of key operations in point cloud algorithms. Delayed-aggregation let point cloud algorithms achieve 1.6$\times$ speedup and 51.1\% energy reduction on a mobile GPU while retaining the accuracy (-0.9\% loss to 1.2\% gains). To maximize the algorithmic benefits, we propose minor extensions to contemporary CNN accelerators, which can be integrated into a mobile Systems-on-a-Chip (SoC) without modifying other SoC components. With additional hardware support, \proj achieves up to 3.6$\times$ speedup.

\end{abstract}

\begin{IEEEkeywords}
Point cloud; DNN; accelerator;
\end{IEEEkeywords}

\begin{Artifact}
\url{https://github.com/horizon-research/efficient-deep-learning-for-point-clouds}
\end{Artifact}


\section{Introduction}
\label{sec:intro}

In recent years, we have seen the explosive rise of intelligent machines that operate on \textit{point clouds}, a fundamental visual data representation that provides a direct 3D measure of object geometry, rather than 2D projections (i.e., images). For instance, Waymo’s self-driving cars carry five LiDAR sensors to gather point clouds from the environment in order to estimate the trajectory over time and to sense object depths~\cite{waymolidar}. Augmented Reality (AR) development frameworks such as Google's ARCore enable processing point clouds for localization (SLAM) and scene understanding~\cite{arcore}. While point cloud algorithms traditionally use ``hand-crafted'' features~\cite{rusu2009fast, tombari2010unique}, they are increasingly moving towards learned features using deep learning~\cite{qi2017pointnet++, wang2019dynamic}, which poses significant efficiency challenges.

We present \proj\footnote{\textipa{[me-s"@ra-z\=e]} Between two vision modes. \textit{meso}-: in the middle; from Ancient Greek \textmu$\acute{\varepsilon}$\textsigma\textomikron\textvarsigma. \textit{orasi}: vision; from Greek $\acute{o}$\textrho\textalpha\textsigma\texteta.}, an algorithm-architecture co-designed system that simultaneously improves the performance and energy efficiency of point cloud algorithms without hurting the accuracy. \proj applies algorithmic and architectural optimizations that exploit characteristics unique to point cloud. Critically, our algorithmic optimizations can directly benefit software running on commodity GPUs without hardware support. Minor augmentations to contemporary DNN accelerators (NPU) unlock further improvements, widening the applicability of \proj.


We start by understanding the characteristics of point cloud algorithms. They inherit the key idea of conventional image/video processing algorithms (e.g., CNNs): extracting features from local windows (neighborhoods) iteratively and hierarchically until the final output is calculated. However, since points in a point cloud are arbitrarily spread in the 3D space, point cloud algorithms require explicit neighbor search and point aggregation operations (as opposed to direct memory indexing) before the actual feature computation.

This leads to two fundamental inefficiencies. First, the three key steps---neighbor search, aggregation, and feature computation---are serialized, leading to long critical path latency. In particular, neighbor search and feature computation dominate the execution time. Second, feature computation operates on aggregated neighbor points, which are inherently redundant representations of the original points, leading to massive memory and computation redundancies.

We propose \textit{delayed-aggregation}, a new algorithmic primitive for building efficient point cloud networks. The key idea is to delay aggregation after feature computation by exploiting the \textit{approximately distributive} property of feature computation over aggregation. In this way, feature computation operates directly on original input points rather aggregated neighbors, significantly reducing the compute cost and memory accesses. In addition, delayed-aggregation breaks the serialized execution chain in existing algorithms, overlapping neighbor search and feature computation---the two performance bottlenecks---to hide long latencies.

To maximize the benefits of delayed-aggregation, we propose minor extensions to conventional DNN accelerators. We find that delayed-aggregation increases the overhead of aggregation, which involves irregular gather operations. The hardware extension co-designs an intelligent data structure partitioning strategy with a small but specialized memory unit to enable efficient aggregation. Our hardware extensions are integrated into generic DNN accelerators without affecting the rest of a mobile Systems-on-a-Chip (SoC).

We evaluate \proj on a set of popular point cloud algorithms and datasets. On the mobile Pascal GPU on TX2, a representative mobile platform today, the delayed-aggregation algorithm alone without hardware support achieves 1.6$\times$ speedup and 51.1\% energy reduction while retaining the accuracy (-0.9\% loss to 1.2\% gains). We implement and synthesize the \proj hardware support in a 16nm process node and integrate it into a state-of-the-art SoC that incorporates a GPU and an NPU. With 3.8\% area overhead to the NPU (\textless 0.05\% of a typical SoC area), \proj achieves up to 3.6$\times$ speedup, which increases to 6.7 on a futuristic SoC with a dedicated neighbor search accelerator.

The artifact is publicly available at \url{https://github.com/horizon-research/efficient-deep-learning-for-point-clouds}. In summary, this paper makes the following contributions:

\begin{itemize}
	\item We comprehensively characterize the performance bottlenecks as well as the memory and compute cost of state-of-the-art point cloud algorithms, and identify the root-causes of the algorithmic inefficiencies.
	\item We propose delayed-aggregation, an efficient algorithm primitive that enables point cloud algorithms to hide the performance bottlenecks and to reduce the overall workload. Delayed-aggregation can readily achieve significant speedup and energy savings on current-generation mobile GPUs without hardware modification.
	\item We co-design hardware with delayed-aggregation to achieve even greater speedups with minor, yet principled, augmentations to conventional DNN accelerators while retaining the modularity of existing SoCs.
\end{itemize}

\section{Background}
\label{sec:bck}

\paragraph{Point Cloud} A point cloud is an unordered set of points in the 3D Cartesian space. Each point is uniquely identified by its $<x$, $y$, $z>$ coordinates. While point cloud has long been used as a fundamental visual data representation in fields such as 3D modeling~\cite{alliez2017culture} and graphics rendering~\cite{gross2011point, levoy1985use, rusinkiewicz2000qsplat, pfister2000surfels}, it has recently received lots of attention in a range of emerging intelligent systems such as autonomous vehicles~\cite{geiger2012we}, robotics~\cite{whitty2010autonomous}, and AR/VR devices~\cite{stets2017visualization}.

\paragraph{Point Cloud Analytics} Similar to conventional visual analytics that analyzes images and videos, point cloud analytics distill semantics information from point clouds. Examples include object detection~\cite{geiger2012we}, semantics segmentation~\cite{behley2019semantickitti}, and classification~\cite{wu20153d}. While image and video analytics have been well-optimized, point cloud analytics require different algorithms and are much less optimized.

Point cloud algorithms operate by iteratively extracting features of each point. Conventional point cloud algorithms use ``hand-crafted'' features such as FPFH~\cite{rusu2009fast} and SHOT~\cite{tombari2010unique}. Recent deep learning-based algorithms use learned features and have generally out-performed conventional algorithms~\cite{chen2016eyeriss}. This paper thus focuses on deep learning-based algorithms.

We focus on deep learning-based algorithms that directly manipulate raw point clouds. Other data representations such as 2D projections of 3D points and voxelization suffer from low accuracy and/or consume excessively high memory~\cite{liu2019point}.

\section{Motivation}
\label{sec:mot}

We first introduce the general flow of point cloud algorithms and identify key operators (\Sect{sec:mot:conv}). We then characterize point cloud algorithms on today's hardware systems to understand the algorithmic and execution bottlenecks (\Sect{sec:mot:char}), which motivate the \proj design.

\begin{figure}[t]
    \centering
    \subfloat[Network architecture of PointNet++~\cite{qi2017pointnet++}.]{
      \label{fig:pointnet2}
      \includegraphics[width=\columnwidth]{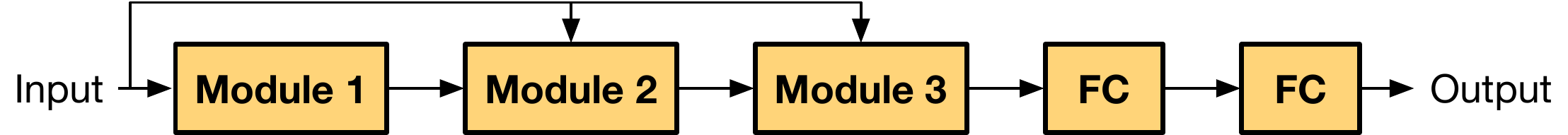}
    }
    \\
    \subfloat[DGCNN~\cite{wang2019dynamic} network architecture. ``\textbf{+}'' is tensor concatenation.]{
      \label{fig:dgcnn}
      \includegraphics[width=\columnwidth]{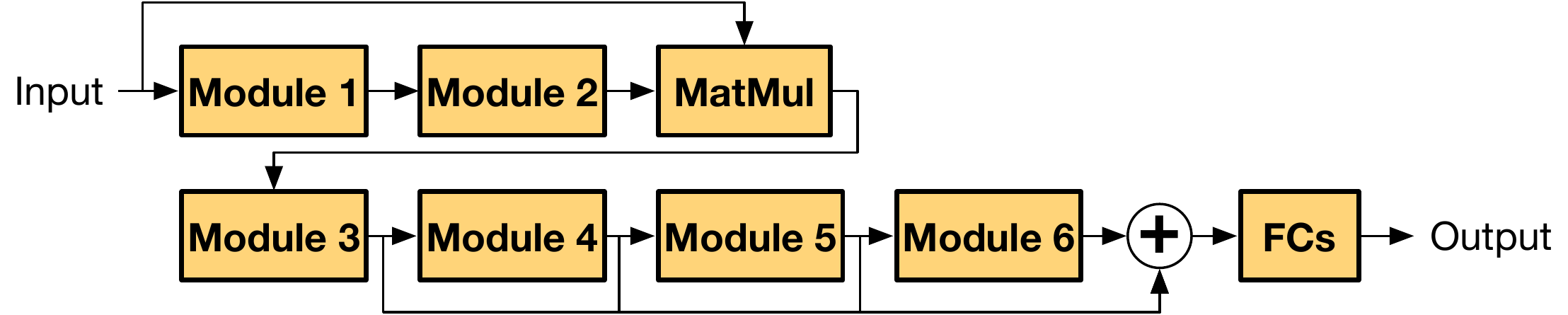}
    }
    \caption{Point cloud networks consist of a set of modules, which extract local features from the input point cloud iteratively and hierarchically to calculate the final output.}
    \label{fig:module}
\end{figure}

\begin{figure}[t]
\centering
\subfloat[Convolution in conventional CNNs can be thought of as two steps: 1) neighbor search ($\cN$) by directly indexing adjacent pixels and 2) feature computation ($\cF$) by a dot product.]
{
  \includegraphics[trim=0 0 0 0, clip, width=\columnwidth]{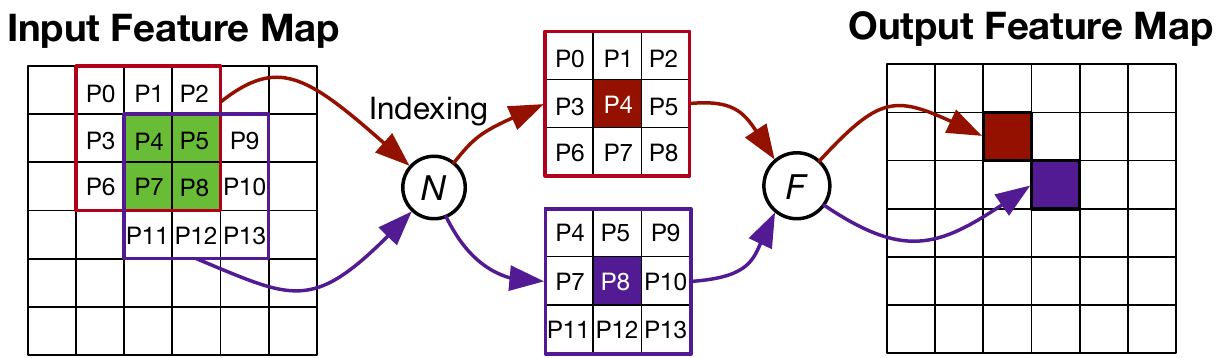}
  \label{fig:convreuse}
}
\\
\subfloat[Point cloud networks consist of three main steps: neighbor search ($\cN$), aggregation ($\cA$), and feature computation ($\cF$). $\cN$ requires an explicit neighbor search; $\cA$ normalizes neighbors to their centroid; $\cF$ is an MLP with batched inputs (i.e., shared MLP weights).]
{
  \includegraphics[trim=0 0 0 0, clip, width=\columnwidth]{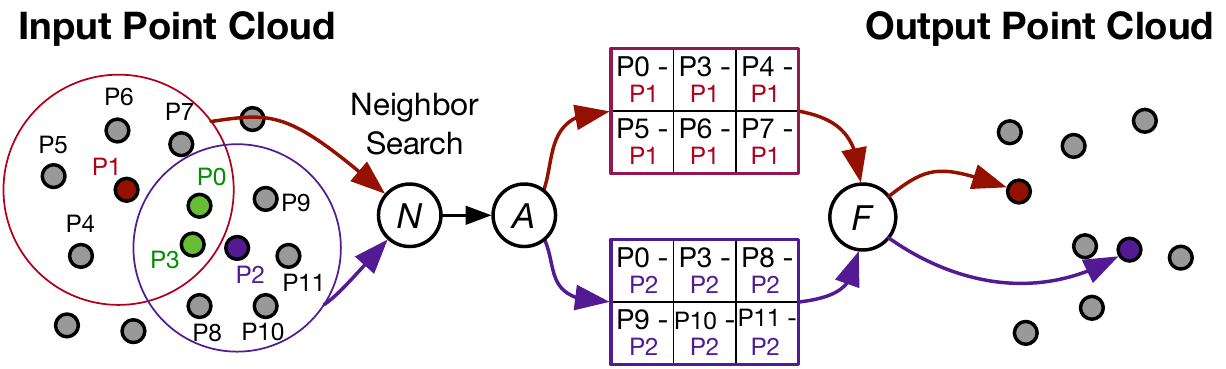}
  \label{fig:nsreuse}
}
\caption{Comparing a convolution layer in conventional CNNs and a module in point cloud networks.}
\label{fig:conv}
\end{figure}

\begin{figure*}[t]
\centering
\includegraphics[width=2.1\columnwidth]{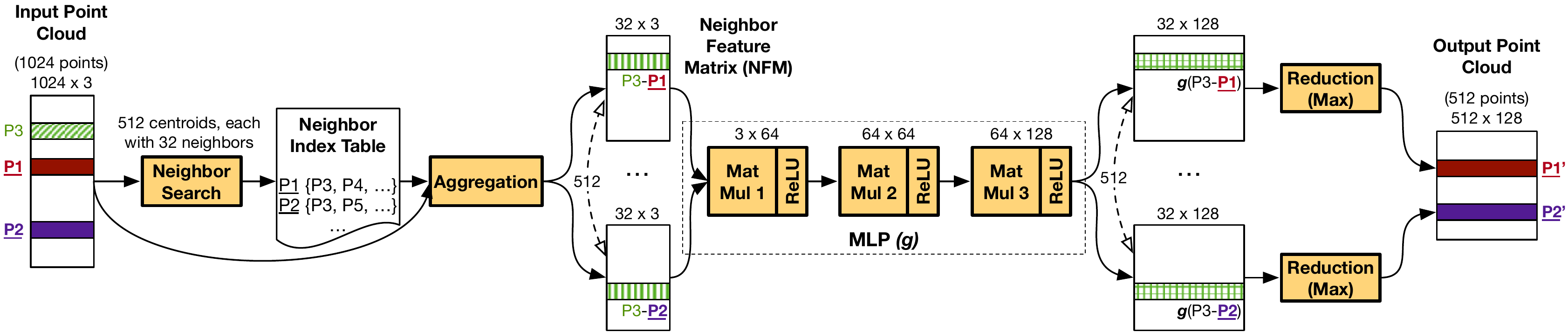}
\caption{The first module in PointNet++~\cite{qi2017pointnet++}. The same MLP is shared across all the row vectors in a Neighbor Feature Matrix (NFM) and also across different NFMs. Thus, MLPs in point cloud networks process batched inputs, effectively performing matrix-matrix multiplications. The (shared) MLP weights are small in size, but the MLP activations are much larger. This is because the same input point is normalized to different values in different neighborhoods before entering the MLP. For instance, \texttt{P3} is normalized to different offsets with respect to \texttt{\underline{P1}} and \texttt{\underline{P2}} as \texttt{P3} is a neighbor of both \texttt{\underline{P1}} and \texttt{\underline{P2}}. In point cloud algorithms, most points are normalized to 20 to 100 centroids, proportionally increasing the MLP activation size.}
\label{fig:baselinealgo}
\end{figure*}

\subsection{Point Cloud Network Architecture}
\label{sec:mot:conv}

\paragraph{Module} The key component in point cloud algorithms is a \textit{module}. Each module transforms an input point cloud to an output point cloud, similar to how a convolution \textit{layer} transforms an input feature map to an output feature map in conventional CNNs. A point cloud network assembles different modules along with other common primitives such as fully-connected (FC) layers. \Fig{fig:pointnet2} and \Fig{fig:dgcnn} illustrate the architecture of two representative point cloud networks, PointNet++~\cite{qi2017pointnet++} and DGCNN~\cite{wang2019dynamic}, respectively.

\paragraph{Module Internals} Each point $\mathbf{p}$ in a point cloud is represented by a feature vector, which in the original point cloud is simply the 3D coordinates of the point. The input point cloud to a module is represented by an $N_{in} \times M_{in}$ matrix, where $N_{in}$ denotes the number of input points and $M_{in}$ denotes the input feature dimension. Similarly, the output point cloud is represented by an $N_{out} \times M_{out}$ matrix, where $N_{out}$ denotes the number of output points and $M_{out}$ denotes the output feature dimension. Note that $N_{in}$ and $N_{out}$ need not be the same; neither do $M_{in}$ and $M_{out}$.



Internally, each module extracts local features from the input point cloud. This is achieved by iteratively operating on a small \textit{neighborhood} of input points, similar to how a convolution layer extracts local features of the input image through a sliding window. \Fig{fig:conv} illustrates this analogy.

Specifically, each output point $\mathbf{p}_o$ is computed from an input point $\mathbf{p}_i$ in three steps --- neighbor search ($\cN$), aggregation ($\cA$), and feature computation ($\cF$):
\begin{align}
  \mathbf{p}_o = \cF(\cA(\cN(\mathbf{p}_i),~~\mathbf{p}_i))
\end{align}
\noindent where $\cN$ returns $K$ neighbors of $\mathbf{p}_i$, $\cA$ aggregates the $K$ neighbors, and $\cF$ operates on the aggregation ($\mathbf{p}_i$ and its $K$ neighbors) to generate the output $\mathbf{p}_o$.

The same formulation applies to the convolution operation in conventional CNNs as well, as illustrated in \Fig{fig:conv}. However, the specifics of the three operations differ in point cloud networks and CNNs. Understanding the differences is key to identifying optimization opportunities.

\paragraph{Neighbor Search} $\cN$ in convolution returns $K$ adjacent pixels in a regular 3D tensor by simply \textit{indexing} the input feature map ($K$ dictated by the convolution kernel volume). In contrast, $\cN$ in point cloud networks requires explicit \textit{neighbor search} to return the $K$ nearest neighbors of $\mathbf{p}_i$, because the points are irregularly scattered in the space. Similar to the notion of a ``stride'' in convolution, the neighbor search might be applied to only a subset of the input points, in which case $N_{out}$ would be smaller than $N_{in}$, as is the case in \Fig{fig:nsreuse}.



\paragraph{Aggregation} Given the $K$ pixels, convolution in CNNs directly operates on the raw pixel values. Thus, conventional convolution skips the aggregation step.

In contrast, point cloud modules operate on the \textit{relative} value of each point in order to correlate a neighbor with its centroid. For instance, a point $\mathbf{p}_3$ could be a neighbor of two centroids $\mathbf{p}_1$ and $\mathbf{p}_2$ (as is the case in \Fig{fig:nsreuse}). To differentiate the different contributions of $\mathbf{p}_3$ to $\mathbf{p}_1$ and $\mathbf{p}_2$, $\mathbf{p}_3$ is \textit{normalized} to the two centroids by calculating the offsets $\mathbf{p}_3 - \mathbf{p}_1$ and $\mathbf{p}_3 - \mathbf{p}_2$ for subsequent computations.

Generally, for each neighbor $\mathbf{p}_k \in \cN(\mathbf{p}_i)$, the aggregation operation calculates the offset $\mathbf{p}_k - \mathbf{p}_i$ (a $1 \times M_{in}$ vector). All $K$ neighbors' offsets form a Neighbor Feature Matrix (NFM) of size $K \times M_{in}$, effectively aggregating the neighbors of $\mathbf{p}_i$.

\begin{figure*}[t]
\centering
\begin{minipage}[t]{0.48\columnwidth}
  \centering
  \includegraphics[trim=0 0 0 0, clip, height=1.40in]{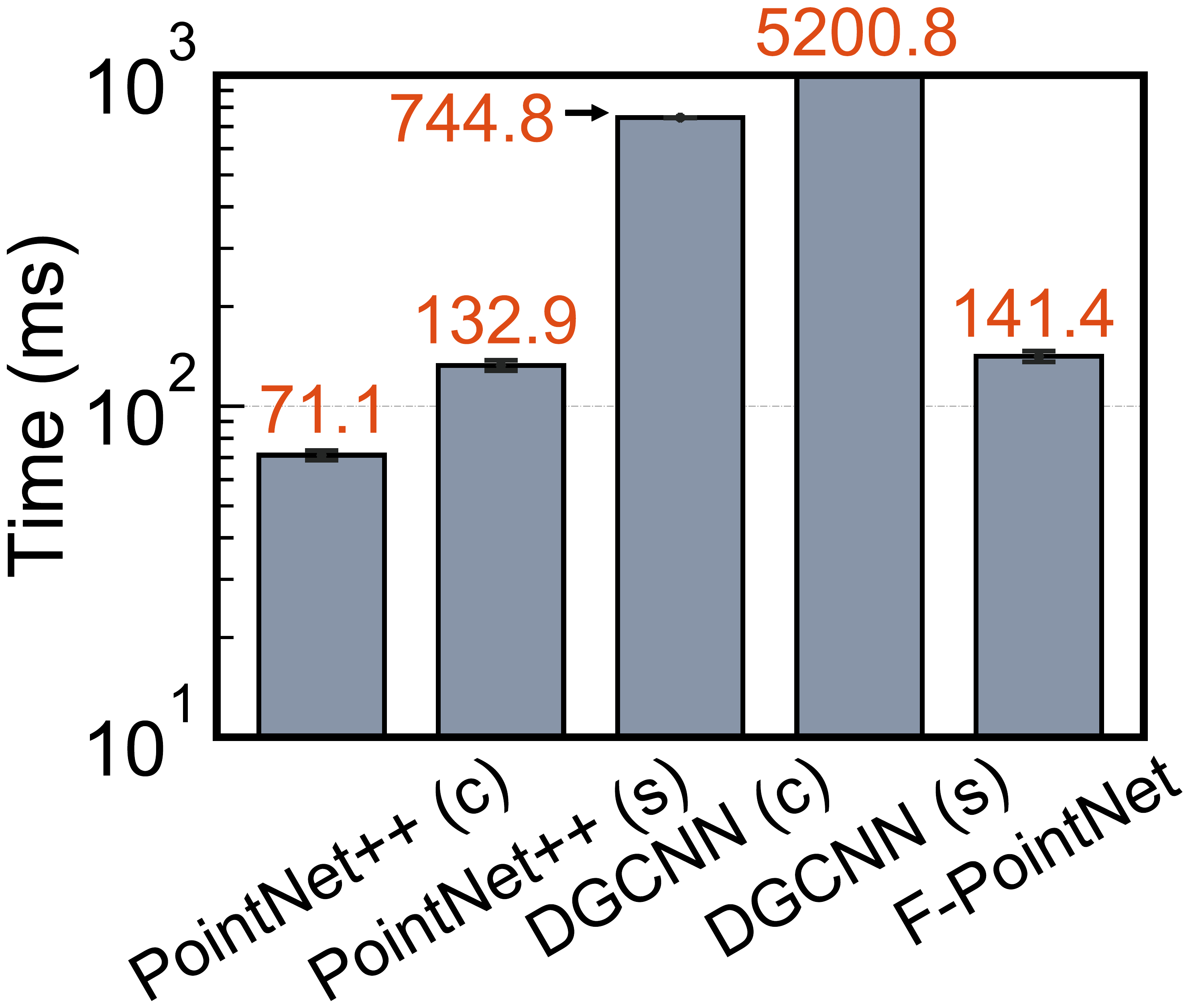}
  \caption{Latency of five point cloud networks on the Pascal GPU on TX2. Results are averaged over 100 executions, and the error bars denote one standard deviation.}
  \label{fig:motivation-exe-time}
\end{minipage}
\hspace{5pt}
\begin{minipage}[t]{0.48\columnwidth}
  \centering
  \includegraphics[trim=0 0 0 0, clip, height=1.40in]{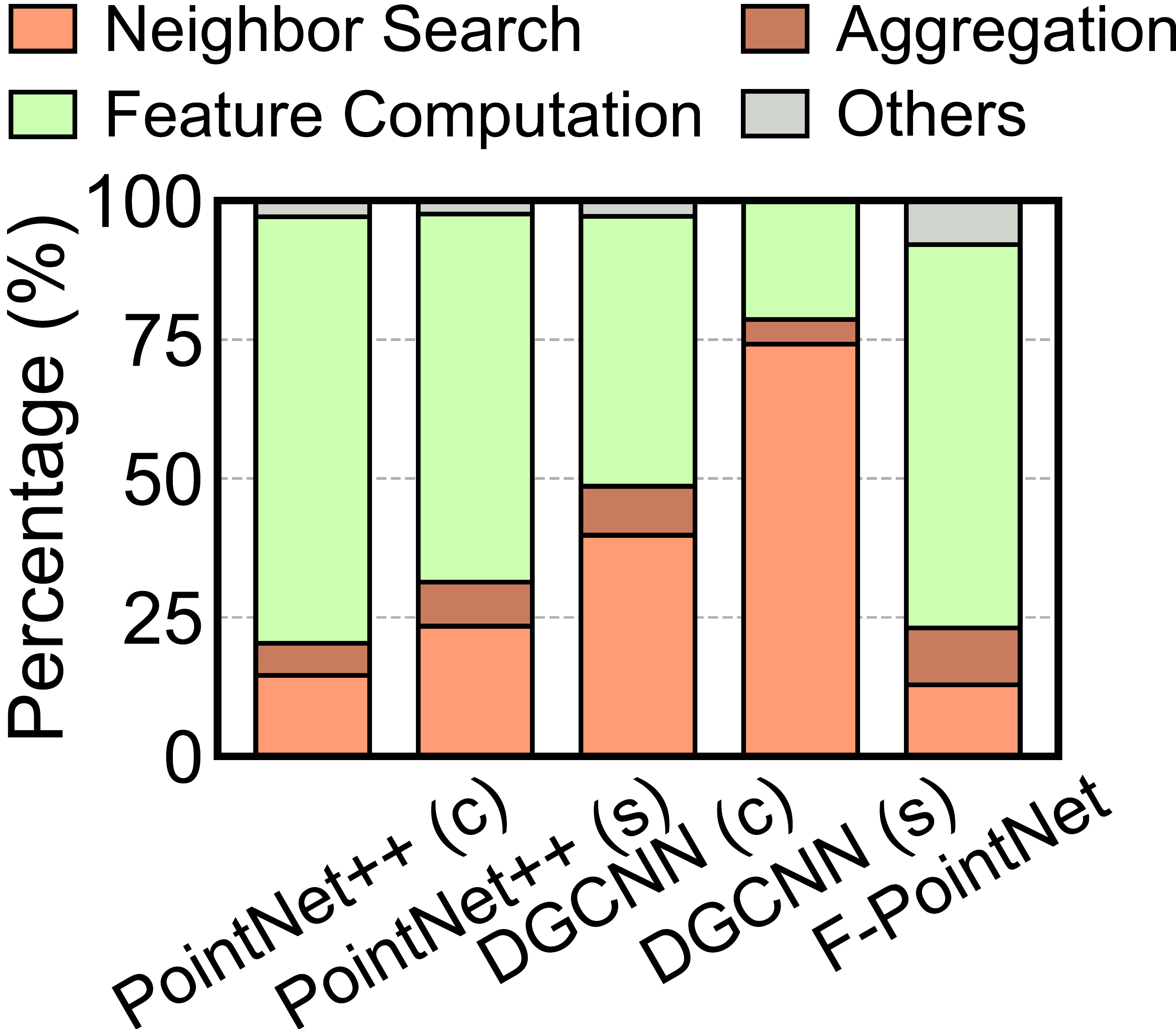}
  \caption{Time distribution across the three main point cloud operations ($\cN$, $\cA$, and $\cF$). The data is averaged on the mobile Pascal GPU on TX2 over 100 executions.}
  \label{fig:motivation-dist}
\end{minipage}
\hspace{5pt}
\begin{minipage}[t]{0.42\columnwidth}
  \centering
  \includegraphics[trim=0 0 0 0, clip, height=1.45in]{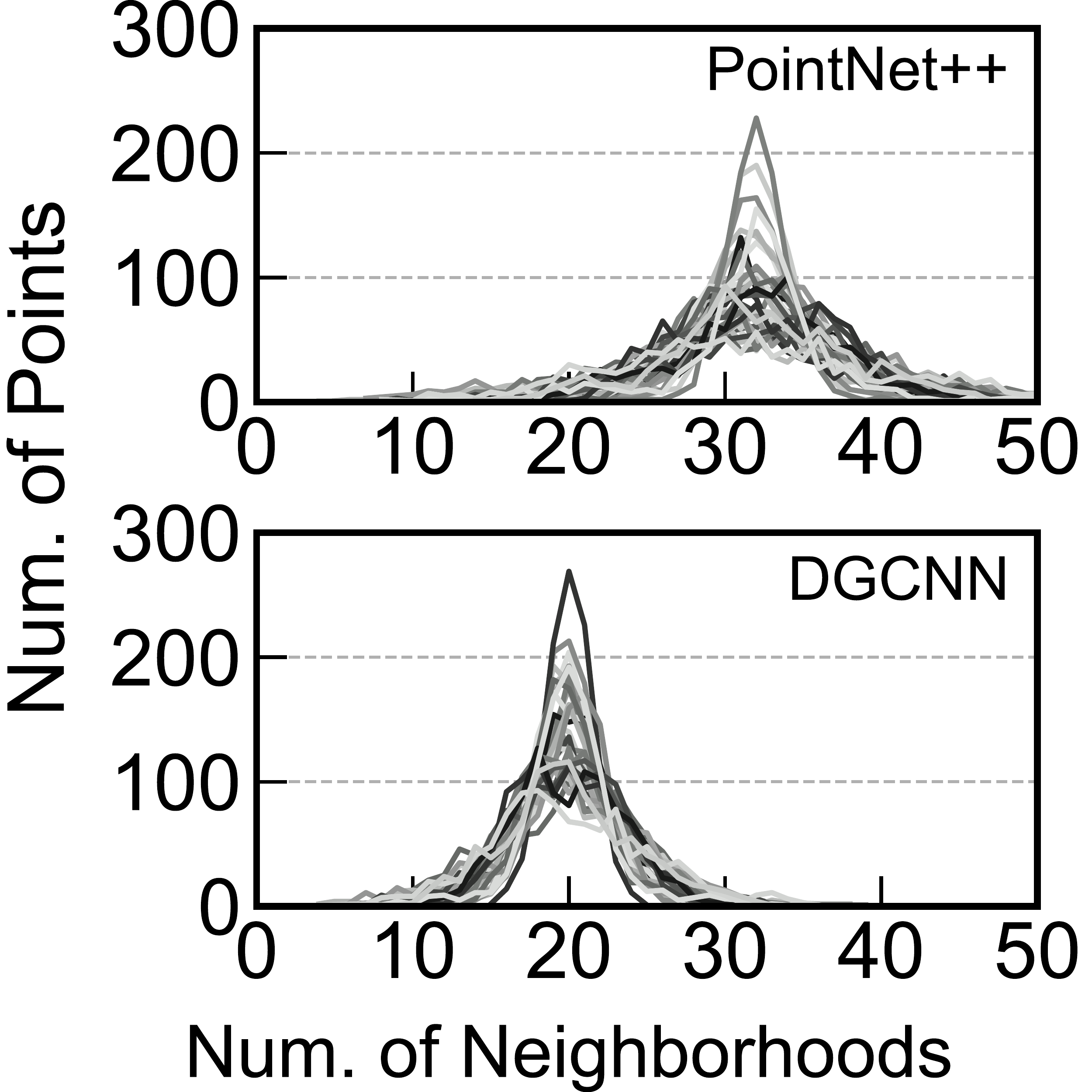}
  \caption{Distribution of the number of points ($y$-axis) that occur in a certain number of neighborhoods ($x$-axis). We profile 32 inputs (curves).}
  \label{fig:motivation-pts-use}
\end{minipage}
\hspace{5pt}
\begin{minipage}[t]{0.55\columnwidth}
  \centering
  \includegraphics[trim=0 0 0 0, clip, height=1.40in]{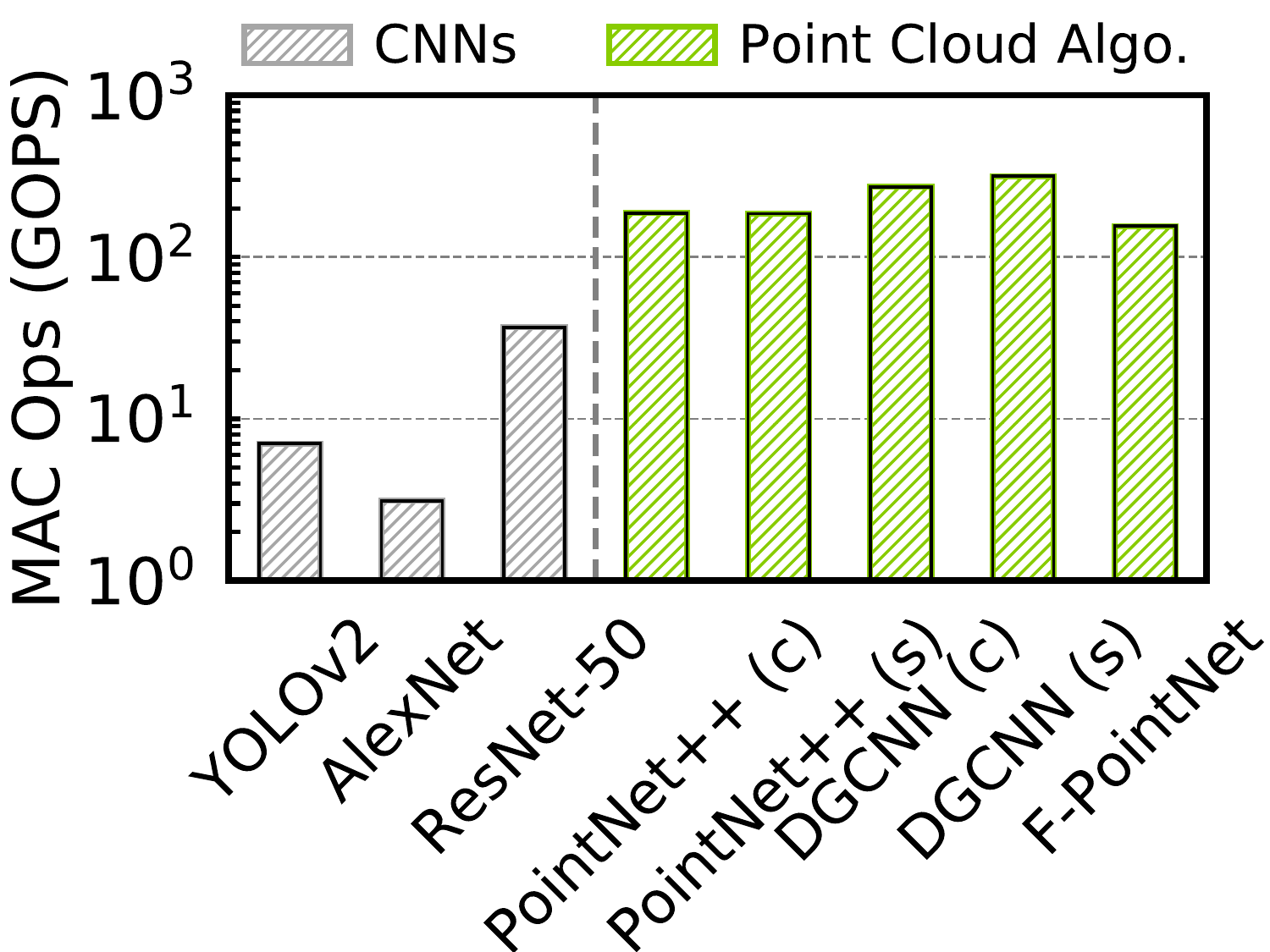}
  \caption{MAC operation comparison between point cloud networks (130K input points per frame~\cite{geiger2012we}) and conventional CNNs (nearly 130K pixels per frame).}
  \label{fig:motivation-mac}
\end{minipage}
\vspace{-5pt}
\end{figure*}

\paragraph{Feature Computation} $\cF$ in convolution is a dot product between the pixel values in a window and the kernel weights. In contrast,  $\cF$ in point cloud applies a multilayer perceptron (MLP) to each row vector in the NFM. Critically, all $K$ row vectors share the same MLP; thus, the $K$ input vectors are batched into a matrix and the MLP becomes a matrix-matrix product, transforming a $K \times M_{in}$ matrix to a $K \times M_{out}$ matrix.

In the end, a reduction operation then reduces the $K \times M_{out}$ matrix to a $1 \times M_{out}$ vector, which becomes the feature vector of an output point. A common choice for reduction is to, for each column independently, take the max of the $K$ rows.

\paragraph{Example} \Fig{fig:baselinealgo} shows the first module in PointNet++~\cite{qi2017pointnet++}, a classic point cloud network that many other networks build upon. This module transforms a point cloud with 1024 ($N_{in}$) points, each with a 3-D ($M_{in}$) feature vector, to a point cloud with 512 ($N_{out}$) points, each with an 128-D ($M_{out}$) feature vector, indicating that the neighbor search is applied to only 512 input points. Each neighbor search returns 32 ($K$) neighbors and forms a $32 \times 3$ NFM, which is processed by a MLP with 3 layers to generate a $32 \times 128$ matrix, which in turn is reduced to a $1 \times 128$ feature vector for an output point. In this particular network, all the NFMs also share the same MLP.

Note that while feature computation is not always MLP and normalization is not always differencing from centroids, they are the most widely used, both in classic networks (e.g., PointNet++~\mbox{\cite{qi2017pointnet++}}) and recent ones (e.g., DGCNN~\mbox{\cite{wang2019dynamic}}).

\subsection{Performance Characterizations}
\label{sec:mot:char}

We characterize point cloud networks on today's systems to understand the bottlenecks and optimization opportunities. To that end, we profile the performance of five popular point cloud networks on the mobile Pascal GPU on the Jetson TX2 development board~\cite{tx2spec}, which is representative of state-of-the-art mobile computing platforms. Please refer to \Sect{sec:exp} for a detailed experimental setup.
 
\paragraph{Time Distribution} \Fig{fig:motivation-exe-time} shows the execution times of the five networks, which range from 71 $ms$ to 5,200 $ms$, clearly infeasible for real-time deployment. The time would scale proportionally as the input size grows.

\Fig{fig:motivation-dist} further decomposes the execution time into the three components, i.e., Neighbor Search ($\cN$), Aggregation ($\cA$), and Feature Computation ($\cF$). $\cN$ and $\cF$ are the major performance bottlenecks. While $\cF$ consists of MLP operations that are well-optimized, $\cN$ (and $\cA$) is uniquely introduced in point cloud networks. Even if $\cF$ could be further accelerated on a DNN accelerator, $\cN$ has compute and data access patterns different from matrix multiplications~\cite{xu2019tigris}, and thus does not fit on a DNN accelerator.

Critically, $\cN$, $\cA$, and $\cF$ are serialized. Thus, they all contribute to the critical path latency; optimizing one alone would not lead to universal speedups. The serialization is inherent to today's point cloud algorithms: in order to extract local features of a point ($\cF$), the point must be aggregated with its neighbors ($\cA$), which in turn requires neighbor search ($\cN$). \proj's algorithm breaks this serialized execution chain, allowing $\cF$ and $\cN$ to be overlapped.

\paragraph{Memory Analysis} Point cloud networks have large memory footprints. While the MLP weights are small and are shared across input NFMs (\Fig{fig:baselinealgo}), the intermediate (inter-layer) activations in the MLP are excessive in size.

\begin{figure*}[t]
\centering
\begin{minipage}[t]{0.65\columnwidth}
  \centering
  \includegraphics[width=\columnwidth]{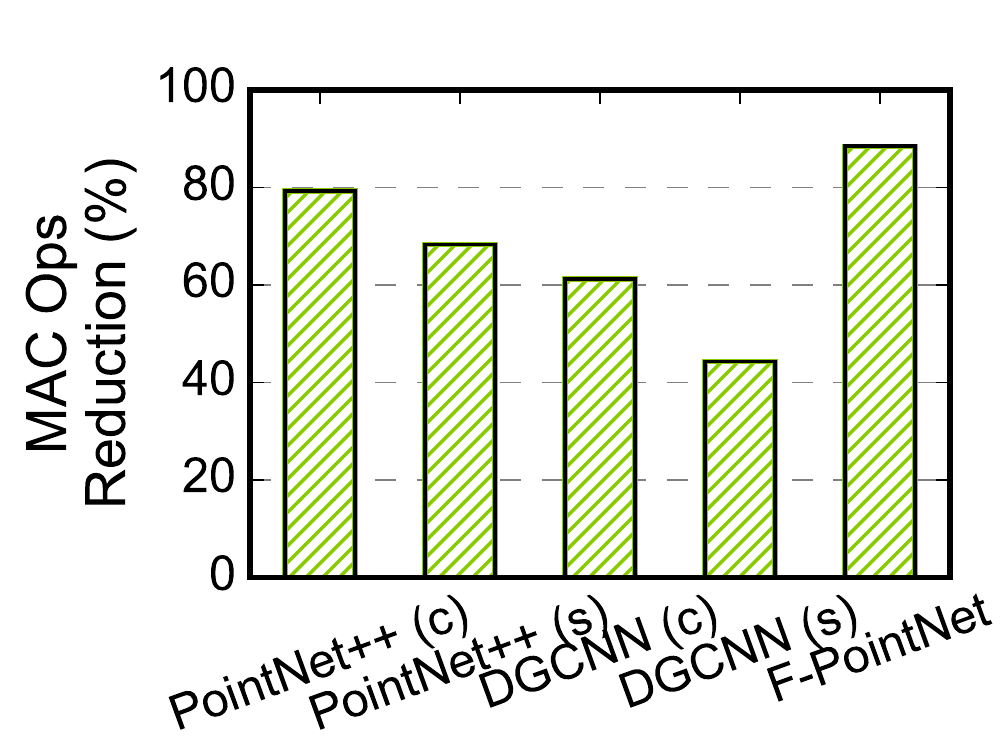}
  \caption{MAC operation reduction in the MLP by delayed-aggregation. The MAC count reductions come from directly operating on the input points as opposed to aggregated neighbors.}
  \label{fig:overall_ops_red}
\end{minipage}
\hspace{5pt}
\begin{minipage}[t]{0.65\columnwidth}
  \centering
  \includegraphics[width=\columnwidth]{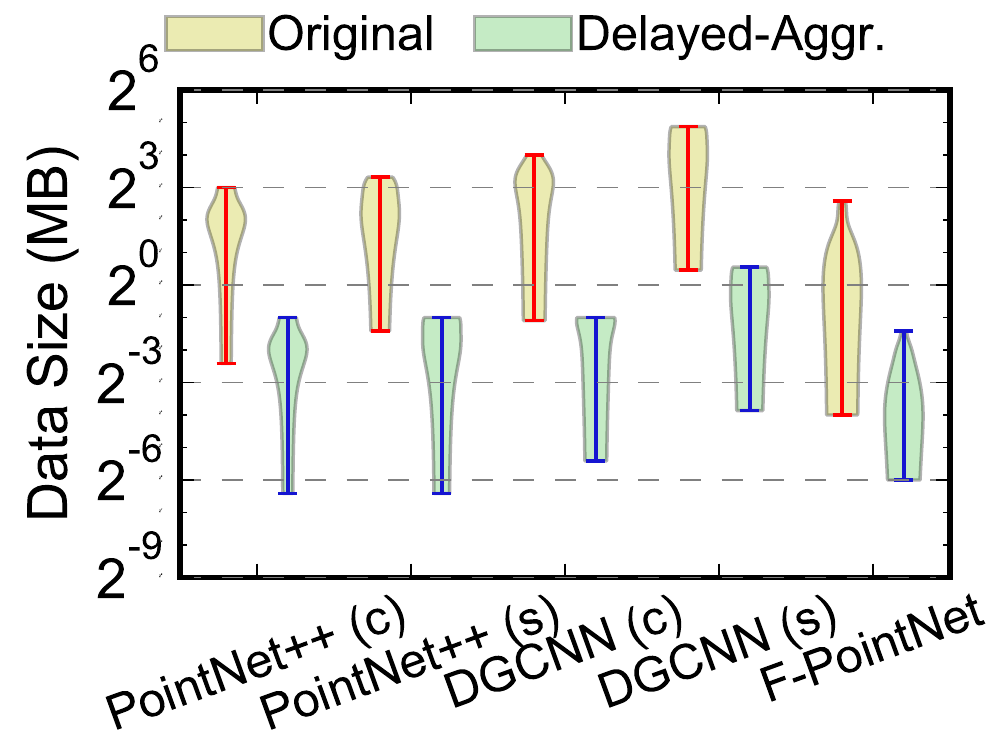}
  \caption{Layer output size distribution as a violin plot with and without delayed-aggregation. High and low ticks denote the largest and smallest layer outputs, respectively.}
  \label{fig:layer_mem_red}
\end{minipage}
\hspace{5pt}
\begin{minipage}[t]{0.65\columnwidth}
  \centering
  \includegraphics[width=\columnwidth]{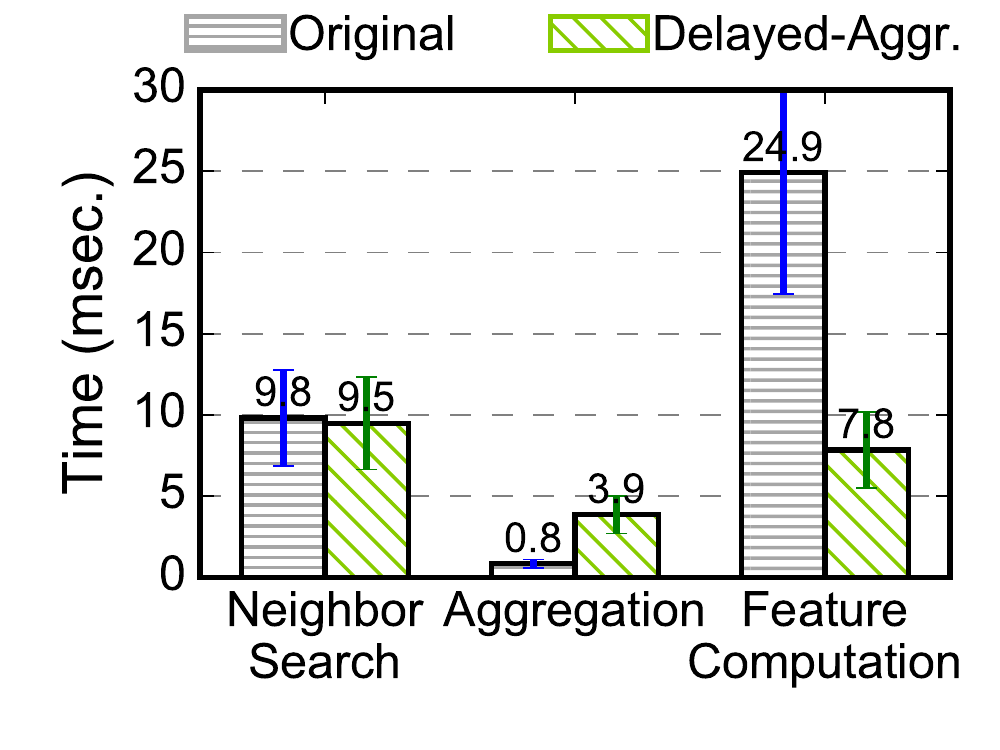}
  \caption{Time distribution across $\cN$, $\cA$, and $\cF$ in PointNet++ (s) with and without delayed-aggregation. Note that delayed-aggregation would also allow $\cN$ and $\cF$ to be executed in parallel.}
  \label{fig:pointnet_dist}
\end{minipage}
\end{figure*}

The ``Original'' category in \Fig{fig:layer_mem_red} shows the distribution of each MLP layer's output size across the five networks. The data is shown as a violin plot, where the high and low ticks represent the largest and smallest layer output size, respectively, and the width of the violin represents the density at a particular size value ($y$-axis). The layer output usually exceeds 2 MB, and could be as large as 32 MB, much greater than a typical on-chip memory size in today's mobile GPUs or DNN accelerators. The large activation sizes would lead to frequent DRAM accesses and high energy consumption.


The large activation size is fundamental to point cloud algorithms. This is because an input point usually belongs to many overlapped neighborhoods, and thus must be normalized to different values, one for each neighborhood. \Fig{fig:nsreuse} shows a concrete example, where \texttt{P3} is a neighbor of both \texttt{P1} and \texttt{P2}; the aggregation operation normalizes \texttt{P3} to \texttt{P1} and \texttt{P2}, leading to two different relative values (\texttt{P3} - \texttt{P1} and \texttt{P3} - \texttt{P2}) that participate in feature computation, increasing the activation size. This is in contrast to convolutions, where pixels in overlapped neighborhoods (windows) are directly reused in feature computation (e.g., \texttt{P4} in \Fig{fig:convreuse}).

\begin{figure*}[t]
\centering
\includegraphics[width=2.1\columnwidth]{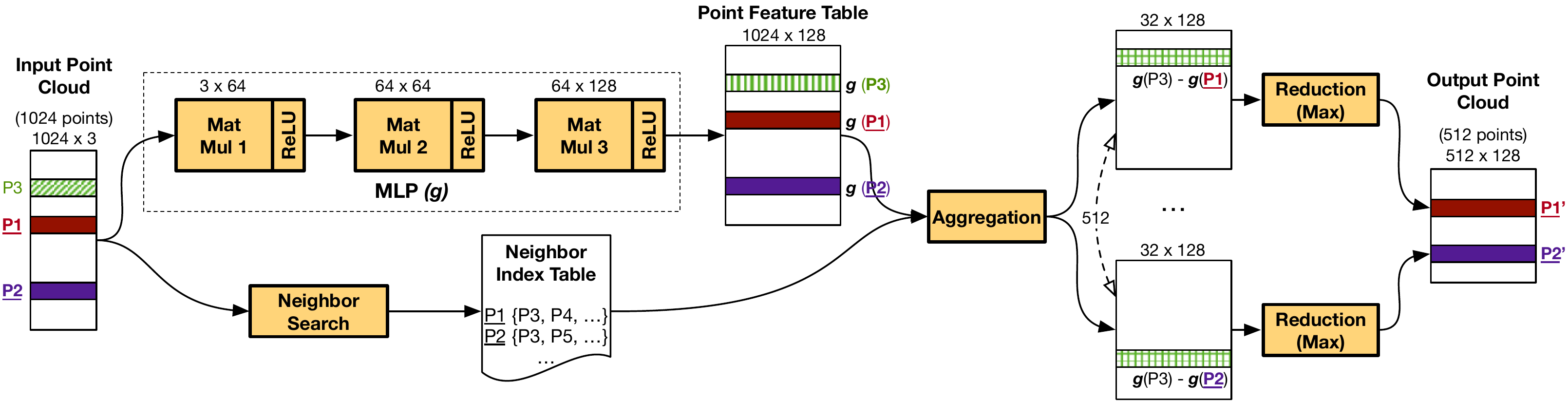}
\caption{The delayed-aggregation algorithm applied to the first module in PointNet++. The MLP and neighbor search are executed in parallel, effectively delaying aggregation after feature computation. The input size of the MLP is much smaller (input point cloud as opposed to the aggregated NFMs), which significantly reduces the MAC operations and the intermediate activation sizes. Aggregation now operates on the output feature space (128-D in this case), whereas it previously operates on the input feature space (3-D in this case). Thus, the aggregation time increases and emerges as a new performance bottleneck.}
\label{fig:newalgo}
\end{figure*}

We use two networks, DGCNN~\cite{qi2017pointnet++} and PointNet++~\cite{ wang2019dynamic}, to explain the large activation sizes. \Fig{fig:motivation-pts-use} shows the distribution of the number of neighborhoods each point is in. Each curve corresponds to an input point cloud, and each ($x$, $y$) point on a curve denotes the number of points ($y$) that occur in a certain number of neighborhoods ($x$). In PointNet++, over half occur in more than 30 neighborhoods; in DGCNN, over half occurs in 20 neighborhoods. Since the same point is normalized to different values in different neighborhoods, this bloats the MLP's intermediate activations.

\paragraph{Compute Cost} The large activations lead to high multiply-and-accumulate (MACs) operations. \Fig{fig:motivation-mac} compares the number of MAC operations in three classic CNNs with that in the feature computation of point cloud networks. To use the same ``resolution'' for a fair comparison, the input point cloud has 130,000 points (e.g., from the widely-used KITTI Odometry dataset~\cite{geiger2012we}) and the CNN input has a similar amount of pixels. In feature computation alone, point cloud networks have an order of magnitude higher MAC counts than conventional CNNs. \proj's algorithm reduces both the memory accesses and MAC counts in feature computation.

\paragraph{Summary} Today's point cloud algorithms extract local features of a point by aggregating the point with its neighbors. The aggregation happens \textit{before} feature computation, which leads to two fundamental inefficiencies:

\begin{itemize}
	\item The two major performance bottlenecks, neighbor search and feature computation, are serialized.
	\item Feature computation operates on aggregated neighbor points, leading to high memory and compute cost.
\end{itemize}

\section{Delayed-Aggregation Algorithm}
\label{sec:algo}

We introduce delayed-aggregation, a  primitive for building efficient point cloud networks (\Sect{sec:algo:algo}). Delayed-aggregation improves the compute and memory efficiencies of point cloud networks without degrading accuracy (\Sect{sec:algo:eff}). We show that aggregation emerges as a new bottleneck in new networks, motivating dedicated hardware support (\Sect{sec:algo:bn}).



\subsection{Algorithm}
\label{sec:algo:algo}


We propose a new framework for building efficient point cloud algorithms. The central idea is to delay aggregation until \textit{after} feature computation so that features are extracted on individual input points rather than on aggregated neighbors. Delayed-aggregation has two benefits. First, it allows neighbor search and feature computation, the two time-consuming components, to be executed in parallel. Second, feature computation operates on input points rather than aggregated neighbors, reducing the compute and memory costs.


\paragraph{Delayed-Aggregation} The key insight is that feature extraction ($\cF$) is \textit{approximately distributive} over aggregation ($\cA$). For an input point $\mathbf{p}_i$ and its corresponding output $\mathbf{p}_o$:

\begin{align}
  \mathbf{p}_o = \cF(\cA(\cN(\mathbf{p}_i),~~\mathbf{p}_i)) \approx \cA(\cF(\cN(\mathbf{p}_i)),~\cF(\mathbf{p}_i))
  \label{eq:fdist}
\end{align}

Fundamentally, \Equ{eq:fdist} holds because the MLP in $\cF$ is approximately distributive over subtraction in $\cA$. Specifically, applying an MLP to the difference of two matrices is approximately equivalent to applying an MLP to both matrices and then subtract the two resulting matrices. The approximation is introduced by the non-linearity in the MLP (e.g., ReLU):

\begin{align}
	& \phi(\phi(
	\begin{bmatrix}
	\mathbf{p}_1 - \mathbf{p}_i \\
	... \\
	\mathbf{p}_k - \mathbf{p}_i
	\end{bmatrix}
	\times W_1) \times W_2) \approx \nonumber \\
	& \phi(\phi(
	\begin{bmatrix}
	\mathbf{p}_1 \\
	... \\
	\mathbf{p}_k
	\end{bmatrix}
	\times W_1 \times W_2)) - \phi(\phi(
	\begin{bmatrix}
	\mathbf{p}_i \\
	... \\
	\mathbf{p}_i
	\end{bmatrix}
	\times W_1 \times W_2))
	\label{eq:mlpdist}
\end{align}

\noindent where $\mathbf{p}_1, ..., \mathbf{p}_k$ are neighbors of $\mathbf{p}_i$, $W_1$ and $W_2$ are the two weight matrices in the MLP (assuming one hidden layer), and $\phi$ is the non-linear activation function. Without $\phi$, the distribution of MLP over subtraction is precise. In actual implementation, the computation on $[\mathbf{p}_i~~...~~\mathbf{p}_i]^\intercal$ is simplified to operating on $\mathbf{p}_i$ once and scattering the result $K$ times.

Critically, applying this distribution allows us to decouple $\cN$ with $\cF$. As shown in \Equ{eq:fdist} and \Equ{eq:mlpdist}, $\cF$ now operates on original input points, i.e., $\mathbf{p}_i$ and $\cN({\mathbf{p}_i})$ (a subset of the input points, too) rather than the normalized point values ($\mathbf{p}_k - \mathbf{p}_i$), which requires neighbor search results. As a result, we could first apply feature computation on all input points. The computed features are then aggregated later.

\paragraph{Walk-Through} We use the first module in PointNet++ as an example to walk through the new algorithm. This module consumes 1024 ($N_{in}$) input points, among which 512 undergo neighbor search. Thus, the module produces 512 ($N_{out}$) output points. The input feature dimension is 3 ($M_{in}$) and the output feature dimension is 128 ($M_{out}$). \Fig{fig:newalgo} shows this module implemented with delayed-aggregation.

We first compute features ($\cF$) from all 1024 points in the input point cloud and store the results in the Point Feature Table (PFT), a $1024 \times 128$ matrix. Every PFT entry contains the feature vector of an input point. Meanwhile, neighbor searches ($\cN$) are executed in parallel on the input point cloud, each returning 32 neighbors of a centroid. The results of neighbor search are stored in a Neighbor Index Table (NIT), a $512 \times 32$ matrix. Each NIT entry contains the neighbor indices of an input point. In the end, the aggregation operation ($\cA$) aggregates features in the PFT using the neighbor information in the NIT. Note that it is the features that are being aggregated, not the original points.

Each aggregated matrix ($32 \times 128$) is reduced to the final feature vector ($1 \times 128$) of an output point. If reduction is implemented by a max operation as is the common case, aggregation could further be delayed after reduction because subtraction is distributive over max: $max(\mathbf{p}_1 - \mathbf{p}_i, \mathbf{p}_2 - \mathbf{p}_i) = max(\mathbf{p}_1, \mathbf{p}_2) - \mathbf{p}_i$. This optimization avoids scattering $\mathbf{p}_i$, reduces the subtraction cost, and is mathematically precise.


\subsection{First-Order Efficiency Analysis}
\label{sec:algo:eff}

Compared with the original implementation of the same module in \Fig{fig:baselinealgo}, the delayed-aggregation algorithm provides three benefits. First, neighbor search and the MLP are now executed in parallel, hiding the latencies of the slower path.

Second, we significantly reduce the MAC operations in the MLP. In this module, the original algorithm executes MLP on 512 $32 \times 3$ matrices while the new algorithm executes MLP only on one $1024 \times 3$ matrix. \Fig{fig:overall_ops_red} shows the MAC operation reductions across all five networks. On average, delayed-aggregation reduces the MAC counts by 68\%.

Third, delayed-aggregation also reduces the memory traffic because the MLP input is much smaller. While the actual memory traffic reduction is tied to the hardware architecture, as a first-order estimation \Fig{fig:layer_mem_red} compares the distribution of per-layer output size with and without delayed-aggregation. The data is shown as a violin plot. Delayed-aggregation reduces the layer output sizes from 8 MB\textasciitilde 32 MB to 512 KB\textasciitilde 1 MB, amenable to be buffered completely on-chip.

By directly extracting features from the input points, our algorithm unlocks the inherent data reuse opportunities in point cloud. Specifically in this example, \texttt{P3} is a neighbor of both \texttt{\underline{P1}} and \texttt{\underline{P2}}, but could not be reused in feature computation by the original algorithm because \texttt{P3}'s normalized values with respect to \texttt{\underline{P1}} and \texttt{\underline{P2}} are different. In contrast, the MLP in our algorithm directly operates on \texttt{P1}, whose feature is then reused in aggregation, \textit{implicitly} reusing \texttt{P1}.

\subsection{Bottleneck Analysis}
\label{sec:algo:bn}

\begin{figure}
    \centering
    \includegraphics[width=\columnwidth]{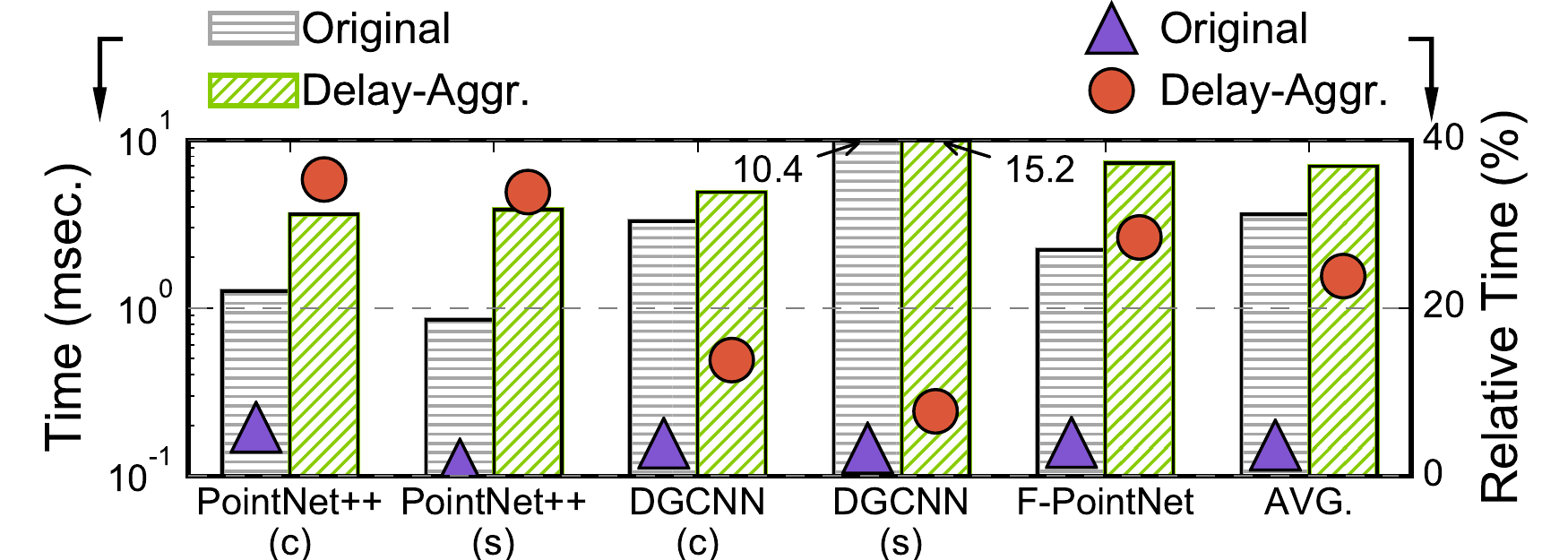}
    \caption{Both absolute (left $y$-axis) and relative (right $y$-axis) aggregation times increase with delayed-aggregation.}
    \label{fig:runtime_dist}
\end{figure}

While delayed-aggregation reduces the compute costs and memory accesses, it also significantly increases the aggregation time. Using PointNet++ as an example, \Fig{fig:pointnet_dist} compares the execution time distribution across the three operations ($\cN$, $\cA$, and $\cF$) with and without delayed-aggregation. The error bars denote one standard deviation in the measurement. The feature extraction time significantly decreases, and the neighbor search time roughly stays the same --- both are expected. The aggregation time, however, significantly increases.

\Fig{fig:runtime_dist} generalizes the conclusion across the five networks. The figure compares the absolute (left $y$-axis) and relative (right $y$-axis) aggregation time in the original and new algorithms. The aggregation time consistently increases in all five networks. Since neighbor search and feature computation are now executed in parallel, aggregation overhead contributes even more significantly to the overall execution time. On average, the aggregation time increases from 3\% to 24\%.

Aggregation time increases mainly because aggregation involves irregular gather operations~\cite{kirk2016programming}, which now operate on a much larger working set with delayed-aggregation. For instance, in PointNet++'s first module (\Fig{fig:newalgo}), aggregation originally gathers from a 12 KB matrix but now gathers from a 512 KB matrix, which is much larger than the L1 cache size (48 KB -- 96 KB\footnote{To our best knowledge, Nvidia does not publish the L1 cache size for the mobile Pascal GPU in TX2 (GP10B~\cite{gp10b}). We estimate the size based on the L1 cache size per SM in other Pascal GPU chips~\cite{pascaluarch} and the number of SMs in the mobile Pascal GPU~\cite{tx2spec}}) in the mobile Pascal GPU on TX2.

The working set size increases significantly because aggregation in new algorithms gathers data from the PFT, whose dimension is $N_{in} \times M_{out}$, whereas the original algorithms gather data from the input point matrix, whose dimension is $N_{in} \times M_{in}$. $M_{out}$ is usually several times greater than $M_{in}$ in order to extract higher-dimensional features. In the example above, $M_{out}$ is 128-D whereas $M_{in}$ is 3-D.

\section{Architectural Support}
\label{sec:arch}


\begin{figure}[t]
\centering
\includegraphics[width=\columnwidth]{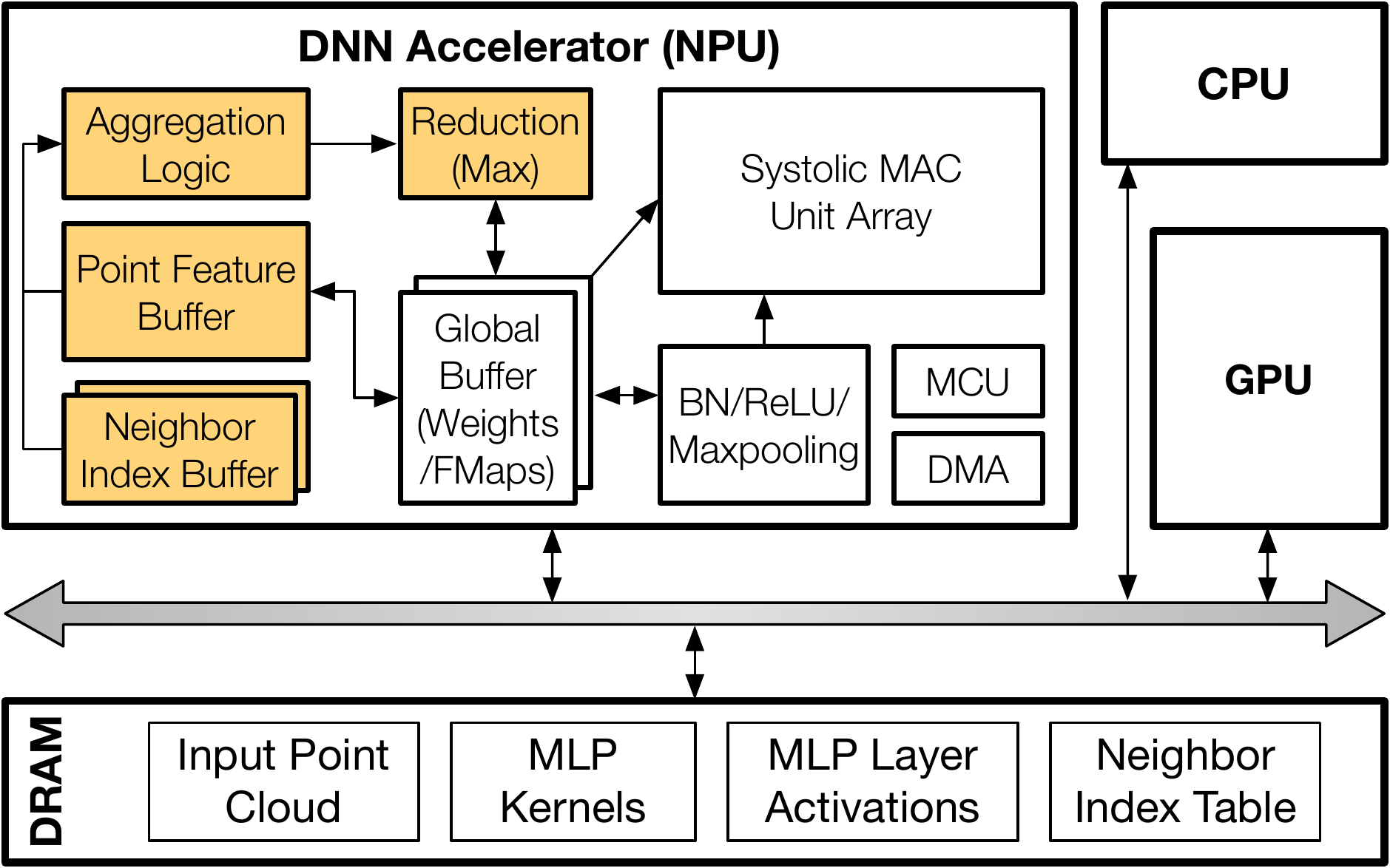}
\caption{The \proj SoC builds on top of today's SoCs consisting of a GPU and a DNN accelerator (NPU). Neighbor search executes on the GPU and feature extraction executes on the NPU. \proj augments the NPU with an aggregation unit (AU) to efficiently execute the aggregation operation. The AU structures are shaded (colored).}
\label{fig:soc}
\end{figure}

\begin{figure*}[t]
\centering
\includegraphics[width=2.1\columnwidth]{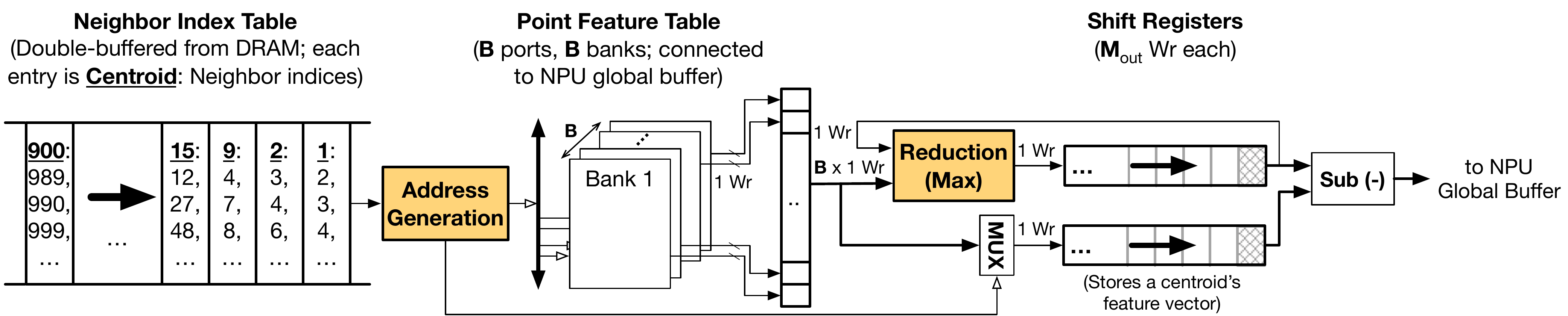}
\caption{Aggregation unit. The NIT buffer is double-buffered from the DRAM. The Address Generation logic simply fetches addresses already buffered in the NIT and sends them to the PFT buffer controller. The PFT buffer is organized as B independently addressed single-ported SRAMs. It could be thought of as an optimized version of a traditional B-banked, B-ported SRAM, because it does not need the crossbar that routes data from banks to ports (but does need the crossbar to route an address to the corresponding bank). The PFT buffer is connected to the NPU's global buffer. Each bank produces one word (Wr) per cycle. The shift registers hold up to $M_{out}$ words, where $M_{out}$ is the output feature vector size. The top shift register holds the result of reduction, and the bottom shift register holds the feature vector of a centroid.}
\label{fig:groupinghw}
\end{figure*}

This section describes \proj, our hardware design that efficiently executes point cloud algorithms developed using delayed-aggregation. \proj extends existing DNN accelerators with minor augmentations while leaving the rest of the SoC untouched. We start from an overview of \proj and its workflow (\Sect{sec:arch:ov}), followed by a detailed description of the architecture support (\Sect{sec:arch:aug}).

\subsection{Overall Design}
\label{sec:arch:ov}

We assume a baseline SoC that incorporates a GPU and an NPU, as with emerging mobile SoCs such as Nvidia Xavier~\cite{xaviersoc}, Apple A13~\cite{applea13}, and Microsoft HPU~\cite{hpuaicore}. Point cloud algorithms are a few times faster when an NPU is available to accelerate MLP compared to running only on the GPU (\Sect{sec:eval:res}). Thus, an NPU-enabled SoC represents the trend of the industry and is a more optimized baseline.

\paragraph{Design} \Fig{fig:soc} shows how \proj augments the NPU in a generic SoC. In \proj, the GPU executes neighbor search ($\cN$) and the NPU executes feature extraction ($\cF$), i.e., the MLP. In addition, \proj augments the NPU with an Aggregation Unit (AU) to efficiently execute the aggregation operation ($\cA$). As shown in \Sect{sec:algo:bn}, aggregation becomes a bottleneck in our new algorithms and is inefficient on the GPU. AU minimally extends a generic NPU architecture with a set of principled memory structures and datapaths.


\proj maps $\cN$ to the GPU because neighbor search is highly parallel, but does not map to the specialized datapath of an NPU. Alternatively, an SoC could use a dedicated neighbor search engine (NSE)~\cite{Kuhara2013An, xu2019tigris}. We use the GPU because it is prevalent in today's SoCs and thus provides a concrete context to describe our design. We later show that delayed-aggregation could achieve even higher speedups in a futurist SoC where an NSE is available to accelerate neighbor search (\Sect{sec:eval:nse}). In either case, \proj does not modify the internals of the GPU or the NSE.




\paragraph{Work Flow} Point cloud algorithms with delayed-aggregation work on \proj as follows. The input point cloud is initially stored in the DRAM. The CPU configures and triggers the GPU and the NPU simultaneously, both of which read the input point cloud. The GPU executes the KNN search and generates the Neighbor Index Table (NIT), which gets stored back to the DRAM. Meanwhile, the NPU computes features for input points and generates the Point Feature Table (PFT). The AU in NPU combines the PFT with the NIT from the memory for aggregation and reduction, and eventually generates the output of the current module.

In some algorithms (e.g., PointNet++), neighbor searches in all modules search in the original 3-D coordinate space, while in other algorithms (e.g., DGCNN) the neighbor search in module $i$ searches in the output feature space of module ($i-1$). In the latter case, the current module's output is written back to the memory for the GPU to read in the next module.



Our design modifies only the NPU while leaving other SoC components untouched. This design maintains the modularity of existing SoCs, broadening the applicability. We now describe the AU augmentation in NPU in detail.




\subsection{Aggregation Unit in NPU}
\label{sec:arch:aug}

Aggregation requires irregular gather operations that are inefficient on GPUs. The key to our architectural support is the specialized memory structures co-designed with customized data structure partitioning, which provide efficient data accesses for aggregation with a little area overhead.

Algorithmically, aggregation iterates over the NIT's $N_{out}$ entries until NIT is exhausted. Each NIT entry contains the $K$ neighbor indices of a centroid $\mathbf{p}$. The aggregation operation first gathers the $K$ corresponding entries (feature vectors) from the PFT ($N_{in} \times M_{out}$). The $K$ feature vectors are then reduced to one ($1 \times M_{out}$) vector, which subtracts $\mathbf{p}$'s feature vector to generate the output feature for $\mathbf{p}$.

\Fig{fig:groupinghw} shows the detailed design of the aggregation unit. The NIT is stored in an SRAM, which is doubled-buffered in order to limit the on-chip memory size. The PFT is stored in a separate on-chip SRAM connected to the NPU's global buffer (which stores the MLP weights and input/output). This allows the output of feature extraction to be directly transferred to the PFT buffer without going through the DRAM. Similarly, the aggregation output is directly written back to the NPU's global buffer, as the aggregation output of the current module is the input to the feature extraction in the next module.

To process each NIT entry, the Address Generation Unit (AGU) uses the $K$ indices to generate $K$ addresses to index into the PFT buffer. Due to the large read bandwidth requirement, the PFT buffer is divided into $B$ independently addressable banks, each of which produces 1 word per cycle.

Each cycle, the PFT buffer produces $B$ words, which enters the reduction unit. In our current design, the reduction unit implements the max logic as is the case in today's point cloud algorithms. The output of the max unit, i.e., the max of the $B$ words, enters a shift register (the top one in \Fig{fig:groupinghw}). Ideally, the number of banks $B$ is the same as the number of neighbors $K$ and the $K$ addresses fall into different banks. If so, the shift register is populated with the $1 \times M_{out}$ vector after $M_{out}$ cycles. The AGU then reads $\mathbf{p}$'s feature vector from the PFT buffer and stores it in another shift register (the bottom one in \Fig{fig:groupinghw}). The two shift registers perform an element-wise subtraction as required by aggregation. The same process continues until the entire NIT is exhausted.

\paragraph{Multi-Round Grouping} In reality, reading the neighbor feature vectors takes more than $M_{out}$ cycles because of two reasons. First, $K$ could be greater than $B$. The number of banks $B$ is limited by the peripheral circuits overhead, which increases as $B$ increases. Second, some of the $K$ addresses could fall into the same bank, causing bank conflicts. We empirically find that an LSB-interleaving reduces bank conflicts, but it is impossible to completely avoid bank conflict at runtime, because the data access patterns in point cloud are irregular and could not be statically calculated -- unlike conventional DNNs and other regular kernels.

We use a simple multi-round design to handle both non-ideal scenarios. Each round the AGU would attempt to identify as many unconflicted addresses as possible, which is achieved by the AGU logic examining each address modulo $B$. The unconflicted addresses are issued to the PFT buffer, whose output enters the max unit to generate a temporary output stored in the shift register. The data in the shift register would be combined with the PFT output in the next round for reduction. This process continues until all the addresses in an NIT buffer entry are processed.

An alternative way to resolve bank-conflict would be to simply ignore conflicted banks, essentially approximating the aggregation operation. We leave it to future work to explore this optimization and its impact on the overall accuracy.


\begin{figure}[t]
\centering
\includegraphics[width=\columnwidth]{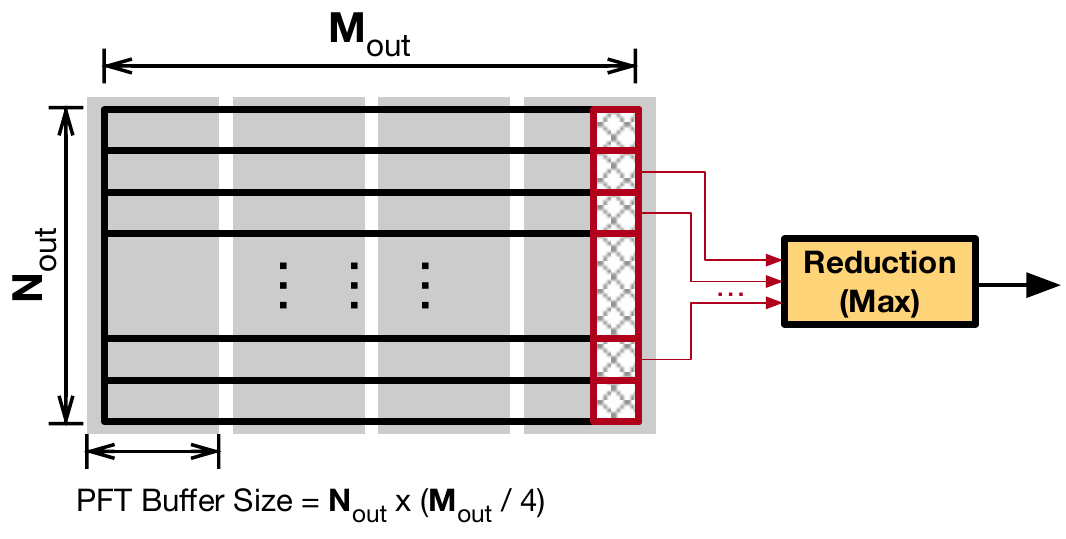}
\caption{Column-major partitioning of PFT to reduce PFT buffer size (4 partitions in this example). Each time the PFT buffer is filled with only one partition. Since reduction (max) is applied to each column independently, the column-major partitioning ensures that all the neighbors of a centroid are present in the PFT buffer for aggregation.}
\label{fig:colpart}
\end{figure}


\paragraph{PFT Buffer Design} One could think of the PFT buffer as a $B$-banked, $B$-ported SRAM. Traditionally, heavily ported and banked SRAMs are area inefficient due to the crossbar that routes each bank's output to the corresponding issuing port~\cite{weste2015cmos}. However, our PFT buffer is much simplified \textit{without} the crossbar. This is by leveraging a key observation that the outputs of all the PFT banks are consumed by the max unit, which executes a \textit{commutative} operation, i.e., $max(a, b)=max(b, a)$. Thus, the output of each bank need not be routed to the issuing port so long as the requested data is correctly produced. This design optimization greatly reduces the area overhead (\Sect{sec:eval:area}).

One might be tempted to reuse the NPU's global buffer for the PFT buffer to save chip area. After all, the PFT \textit{is} MLP's output, which is stored in the global buffer. However, physically sharing the two SRAM structures is difficult, mainly because of their different design requirements. Global buffer contains MLP weights and layer inputs, accesses to which have regular patterns. As a result, NPU global buffers are usually designed with very few ports (e.g., one)~\cite{jouppi2017datacenter, armmlproc} while using a wide word. In contrast, accesses to the PFT are irregular as the neighbors of a centroid could be arbitrary spread in the PFT. Thus, the PFT buffer must be heavily-ported in order to sustain a high bandwidth requirement.

\paragraph{PFT Partitioning} To hold the entire PFT, the buffer must hold $N_{out} \times M_{out}$ features, which could be as large as 0.75~MB in some networks (e.g., DGCNN). Since the PFT buffer adds area overhead, we would like to minimize its size.

We partition the PFT to reduce the PFT buffer size. Each time, the PFT buffer is filled with only one partition. One straightforward strategy is the row-major partitioning, where the PFT buffer holds only a few rows of the PFT. However, since a centroid's neighbors could be arbitrarily spread across different PFT rows, row-major partitioning does not guarantee that all the neighbors of a centroid are present in the PFT buffer (i.e., in the same partition) for aggregation.

Instead, our design partitions the PFT column-wise, where each partition contains several columns of the PFT. \Fig{fig:colpart} illustrates the idea with 4 partitions. In this way, aggregation of a centroid is divided into four steps, each step aggregating only one partition. The column-major partitioning ensures that, within each partition, the neighbors of a centroid are available in the PFT buffer. Since reductions (max) of different columns are independent, the four intermediate reduction results can simply be concatenated in the end.


With column-wise partitioning, each NIT entry is accessed multiples times---once per aggregation step. Thus, a smaller PFT buffer, while reducing the area overhead, would also increase the energy overhead. We later quantify this resource vs. energy trade-off (\Sect{sec:eval:sen}).

%



\section{Experimental Setup}
\label{sec:exp}

%

\paragraph{Hardware Implementation} We develop RTL implementations for the NPU and its augmentations for the aggregation unit (AU). The NPU is based on the systolic array architecture, and consists of a $16 \times 16$ PE array. Each PE consists of two input registers, a MAC unit with an accumulator register, and simple trivial control logic. This is identical to the PE in the TPU~\cite{jouppi2017datacenter}. Recall that MLPs in point cloud networks process batched inputs (\Fig{fig:baselinealgo}), so the MLPs perform matrix-matrix product that can be efficiently implemented on a systolic array. The NPU's global buffer is sized at 1.5 MB and is banked at a 128 KB granularity.

The PFT buffer in the AU is sized at 64 KB with 32 banks. The NIT buffer is doubled-buffered; each buffer is implemented as one SRAM bank sized at 12 KB and holds 128 entries. The NIT buffer produces one entry per cycle. Each entry is 98 Bytes, accommodating 64 neighbor indices (12 bits each). Each of the two shift registers is implemented as 256 flip-flops (4-byte each). The datapath mainly consists of 1) one 33-input max unit and 256 subtraction units in the reduction unit, and 2) 32 32-input MUXes in the AGU.

The design is clocked at 1 GHz. The RTL is implemented using Synposys synthesis and Cadence layout tools in TSMC 16nm FinFET technology, with SRAMs generated by an Arm memory compiler. Power is simulated using Synopsys PrimeTimePX, with fully annotated switching activity.

\paragraph{Experimental Methodology} The latency and energy of the NPU (and its augmentation) are obtained from post-synthesis results of the RTL design. We model the GPU after the Pascal mobile GPU in the Nvidia Parker SoC hosted on the Jetson TX2 development board~\cite{tx2spec}. The SoC is fabricated in a 16 nm technology node, same as our NPU. We directly measure the GPU execution time as well as the kernel launch time. The GPU energy is directly measured using the built-in power sensing circuity on TX2.

The DRAM parameters are modeled after Micron 16 Gb LPDDR3-1600 (4 channels) according to its datasheet~\cite{micronlpddr3}. DRAM energy is calculated using Micron's System Power Calculators~\cite{microdrampower} using the memory traffic, which includes: 1) GPU reading input point cloud, 2) NPU accessing MLP kernels and activations each layer, and 3) GPU writing NIT and NPU reading NIT. Overall, the DRAM energy per bit is about 70$\times$ of that of SRAM, matching prior work~\cite{Yazdanbakhsh2018GAN, gao2017tetris}.

The system energy is the aggregation of GPU, NPU, and DRAM. The overall latency is sum of GPU, NPU, and DRAM minus: 1) double buffering in the NPU, and 2) parallel execution between neighbor search on GPU and feature computation on NPU. Due to double-buffering, the overall latency is dominated by the compute latency, not memory.




\begin{table} 
\caption{Evaluation benchmarks.}
\resizebox{\columnwidth}{!}{
\renewcommand*{\arraystretch}{1}
\renewcommand*{\tabcolsep}{10pt}
\begin{tabular}{ cccc } 
\toprule[0.15em]
\textbf{\specialcell{Application\\Domains}} & \textbf{\specialcell{Algorithm}} & \textbf{Dataset} & \textbf{Year} \\ 
\midrule[0.05em]
\specialcell{Classification} &  \specialcell{PointNet++ (c)\\DGCNN (c)\\LDGCNN\\DensePoint} & ModelNet40 & \specialcell{2017\\2019\\2019\\2019} \\
\midrule[0.05em]
\specialcell{Segmentation} &  
\specialcell{PointNet++ (s)\\DGCNN (s)} & ShapeNet & \specialcell{2017\\2019} \\
\midrule[0.05em]
\specialcell{Detection} &  \specialcell{F-PointNet} & KITTI & \specialcell{2018} \\
\bottomrule[0.15em]
\end{tabular}
}
\label{tab:eval_app}
\end{table}

\paragraph{Software Setup} \Tbl{tab:eval_app} lists the point cloud networks we use, which cover different domains for point cloud analytics including object classification, segmentation, and detection. The networks cover both classic and recent ones (2019).

For classification, we use four networks: PointNet++~\cite{qi2017pointnet++}, DGCNN~\cite{wang2019dynamic}, LDGCNN~\cite{zhang2019linked}, and DensePoint~\cite{liu2019densepoint}; we use the ModelNet40~\cite{wu20153d} dataset. We report the standard overall accuracy metric. To evaluate segmentation, we use the variants of PointNet++ and DGCNN specifically built for segmentation, and use the ShapeNet dataset~\cite{shapenet2015}. We report the standard mean Intersection-over-Unit (mIoU) accuracy metric. Finally, we use F-PointNet~\cite{qi2018frustum} as the object detection network. We use the KITTI dataset~\cite{geiger2012we} and report the geometric mean of the IoU metric (BEV) across its classes.


We optimize the author-released open-source version of these networks to obtain stronger software baselines. We: 1) removed redundant data duplications introduced by \texttt{tf.tile}; 2) accelerated the CPU implementation of an important kernel, 3D Interpretation (\texttt{three\_interpolate}), with a GPU implementation; 3) replaced the Farthest Point Sampling with random sampling in PointNet++ with little accuracy loss; 4) replaced the Grouping operation (\texttt{group\_point}) with an optimized implementation (\texttt{tf.gather}) to improve the efficiency of grouping/aggregation. On TX2, our baseline networks are 2.2$\times$ faster than the open-source versions.

\paragraph{Baseline} We mainly compare with a generic NPU+GPU SoC without any \proj-related optimizations. Compared to the baseline, our proposal improves both the software, i.e., the delayed-aggregation algorithm as well as hardware, i.e., the aggregation unit (AU) augmentations to the NPU.





\paragraph{Variants} To decouple the contributions of our algorithm and hardware, we present two different \proj variants:
\begin{itemize}
    \item \sys{\proj-SW}: delayed-aggregation without AU support. Neighbor search and aggregation execute on the GPU; feature computation executes on the NPU.
    \item \sys{\proj-HW}: delayed-aggregation with AU support. Neighbor search executes on the GPU; aggregation and feature computation execute on the NPU.
\end{itemize}

\section{Evaluation}
\label{sec:eval}

We first show \proj adds little hardware overhead (\Sect{sec:eval:area}) while achieving comparable accuracy against original point cloud networks (\Sect{sec:eval:acc}). We then demonstrate the efficiency gains of \proj on different hardware platforms (\Sect{sec:eval:gpu} -- \Sect{sec:eval:nse}), followed by sensitivity studies (\Sect{sec:eval:sen}).

\subsection{Area Overhead}
\label{sec:eval:area}

\proj introduces only minimal area overhead with the minor AU augmentations. The main overhead comes from the 88 KB additional SRAM required for the PFT buffer and the NIT buffer. Compared to the baseline NPU, the additional hardware introduces less than 3.8\% area overhead (\SI{0.059}{\mm\squared}), which is even more negligible compared to the entire SoC area (e.g., \SI{350}{\mm\squared} for Nvidia Xavier~\cite{xaviersochotchips} and \SI{99}{\mm\squared} for Apple A13~\cite{applea13}).

Our custom-designed PFT buffer avoids the crossbar connecting the banks to the read ports by exploiting the algorithmic characteristics (\Sect{sec:arch:aug}). Since our PFT buffer is heavily banked (32) and ported (32) and each bank is small in size (2 KB), the additional area overhead introduced by the crossbar would have been high. Specifically, the area of the PFT buffer now is \SI{0.031}{\mm\squared}, but the crossbar area would be \SI{0.064}{\mm\squared}, which is now avoided.

\subsection{Accuracy}
\label{sec:eval:acc}

\begin{figure}[t]
  \centering
  \includegraphics[width=\columnwidth]{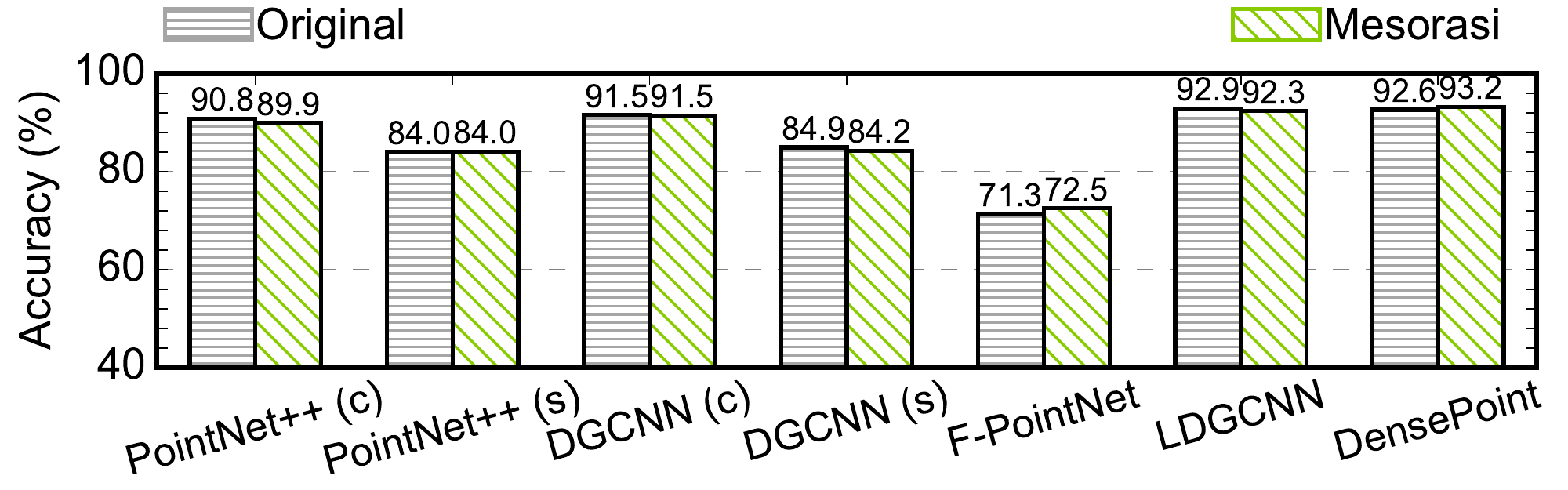}
  \caption{The accuracy comparison between networks trained with delayed-aggregation and the original networks.}
  \label{fig:app_acc}
\end{figure}

Overall, \proj matches or out-performs the original algorithms. We train all seven networks with delayed-aggregation from scratch until the accuracy converges. \Fig{fig:app_acc} compares our accuracy with that of the baseline models, which we choose the better of the reported accuracies in the original papers or accuracies from training their released code. Overall, \proj leads to at most 0.9\% accuracy loss in the case of PointNet++ (c) and up to 1.2\% accuracy gain in the case of F-PointNet. This shows that, while delayed-aggregation approximates the original algorithms, the accuracy loss could be recovered from training. Delayed-aggregation could be used as a primitive to build accurate point cloud algorithms.

We find that fine-tuning the model weights trained on the original networks has similar accuracies as retraining from scratch. However, directly using the original weights without retraining leads to a few percentages of accuracy loss, which is more significant when the non-linear layers use batch normalization, which perturbs the distributive property of matrix multiplication over subtraction more than ReLU.

\subsection{Results on GPU}
\label{sec:eval:gpu}

\begin{figure}
    \centering    \includegraphics[width=\columnwidth]{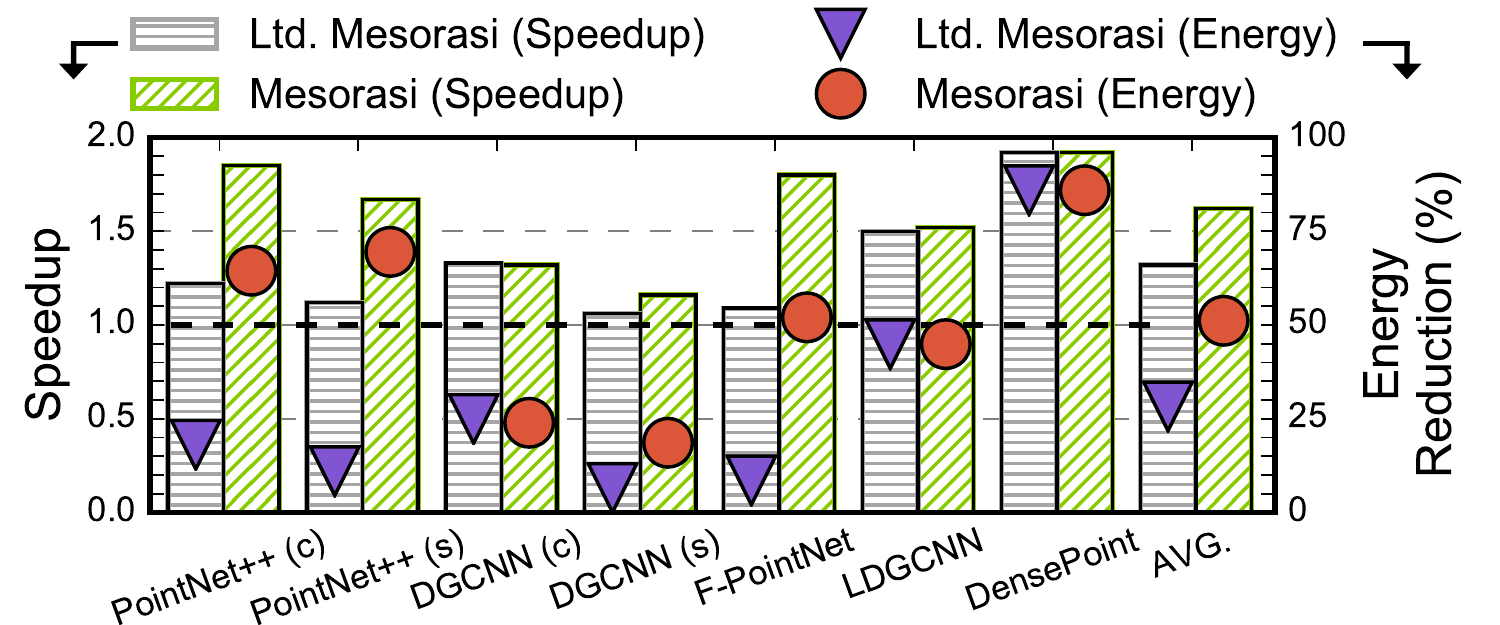}
    \caption{Speedup and energy reduction of the delayed-aggregation algorithm and the limited version of the algorithm on the mobile Pascal GPU on TX2.}
    \label{fig:sw_speedup_energy}
\end{figure}

We first show that our delayed-aggregation algorithm readily achieves significant speedups on today's GPU \textit{without} hardware support. \Fig{fig:sw_speedup_energy} shows the speedup and energy reduction of \proj on the Pascal GPU on TX2.

As a comparison, we also show the results of a limited version of delayed-aggregation, where only the matrix-vector multiplication (MVM) part of an MLP is hoisted before aggregation (Ltd-\proj). The limited delayed-aggregation algorithm is inspired by certain Graph Neural Network (GNN) implementations such as GCN~\cite{gcnpytorch, fey2019pytorchgeo, wang2019deep} and GraphSage~\cite{graphsagetf}. Note that by hoisting only the MVM rather than the entire MLP, Ltd-\proj is precise since MVM is linear. We refer interested readers to the wiki page of our code repository~\cite{mesorasiwiki} for a detailed comparison between our delayed-aggregation and GNN's limited delayed-aggregation.

On average, \proj achieves $1.6\times$ speedup and 51.1\% energy reduction compared to the original algorithms. In comparison, the limited delayed-aggregation algorithm achieves only $1.3\times$ speedup and 28.3\% energy reduction. Directly comparing with Ltd-\proj, \proj has $1.3\times$ speedup and 25.9\% energy reduction. This is because the limited delay-aggregation, in order to be precise, could be applied to only the first MLP layer. By being approximate, \proj does not have this constraint and thus enables larger benefits; the accuracy loss could be recovered through fine-tuning (\Fig{fig:app_acc}). \proj has similar performance as Ltd-\proj on DGCNN (c), LDGCNN, and DensePoint, because these three networks have only one MLP layer per module.


Although delayed-aggregation allows neighbor search and feature extraction to be executed in parallel, and our implementation does exploit the concurrent kernel execution in CUDA, we find that neighbor search and feature extraction in actual executions are rarely overlapped. Further investigation shows that this is because the available resources on the Pascal GPU on TX2 are not sufficient to allow both kernels to execute concurrently. We expect the speedup to be even higher on more powerful mobile GPUs in the future.

Overall, networks in which feature computation contributes more heavily to the overall time, such as PointNet++ (c) and F-PointNet (\Fig{fig:motivation-dist}), have higher MAC operation reductions (\Fig{fig:overall_ops_red}), and thus have higher speedups and energy reductions. This confirms that the improvements are mainly attributed to optimizing the MLPs in feature computation.



\subsection{Speedup and Energy Reduction}
\label{sec:eval:res}

\begin{figure}[t]
\centering
\subfloat[Speedup. Higher is better.]{
	\label{fig:hw_speedup}	\includegraphics[width=\columnwidth]{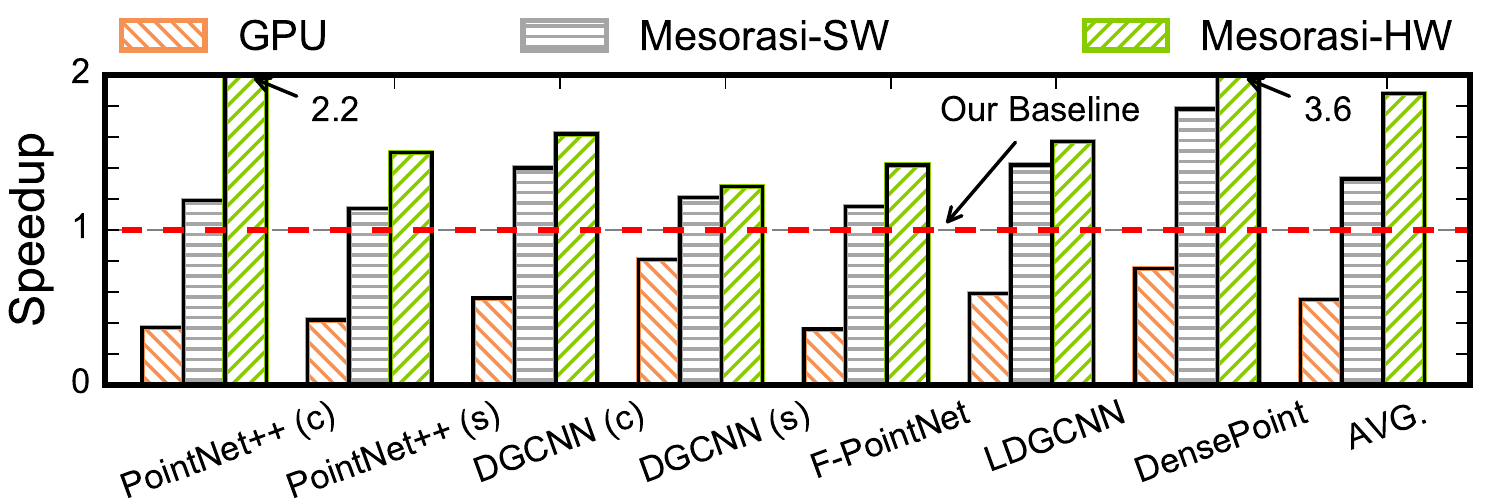} }
\\
\vspace{-5pt}
\subfloat[Normalized energy. Lower is better.]{
	\label{fig:hw_energy}
	\includegraphics[width=\columnwidth]{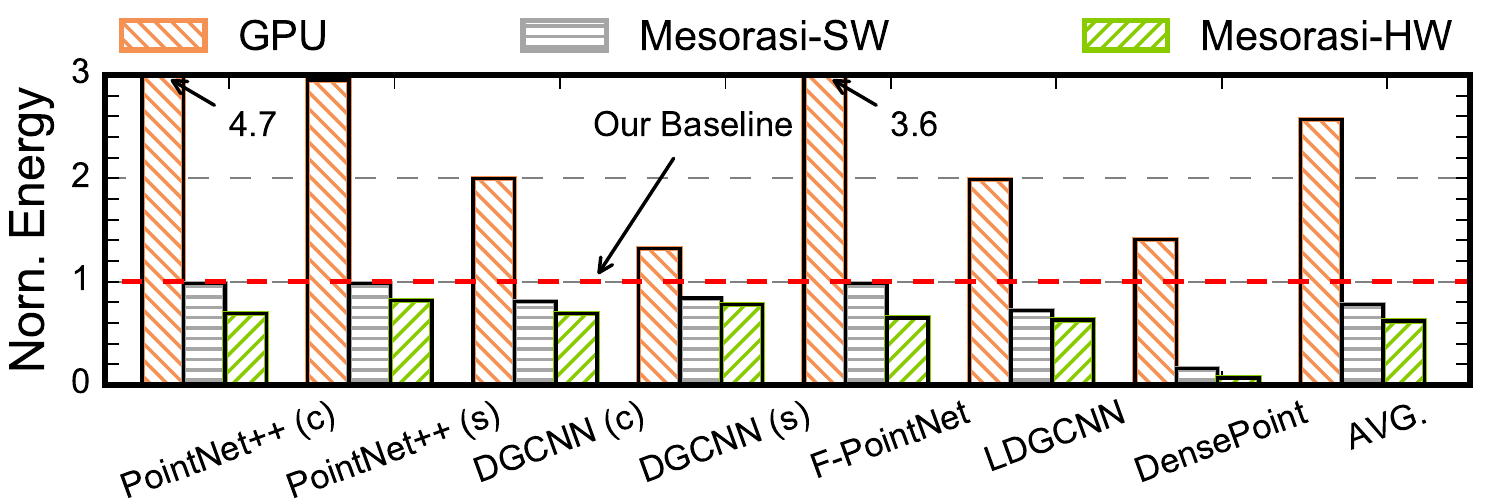} } 
\caption{Speedup and energy reduction of \sys{\proj-SW} and \sys{\proj-HW} over the baseline (GPU+NPU), which is twice as fast and consumes one-third of the energy compared to the GPU, indicating an optimized baseline to begin with.}
\label{fig:hw_speedup_energy}
\end{figure}

\proj also improves the performance and energy consumption of emerging mobile SoCs with a dedicated NPU. \Fig{fig:hw_speedup} and \Fig{fig:hw_energy} show the speedup and the normalized energy consumption of the two \proj variants over the NPU+GPU baseline, respectively.

\begin{figure}[t]
  \centering
  \captionsetup[subfigure]{width=0.5\columnwidth}
  \subfloat[\small{Feature computation.}]
  {
  \includegraphics[width=.48\columnwidth]{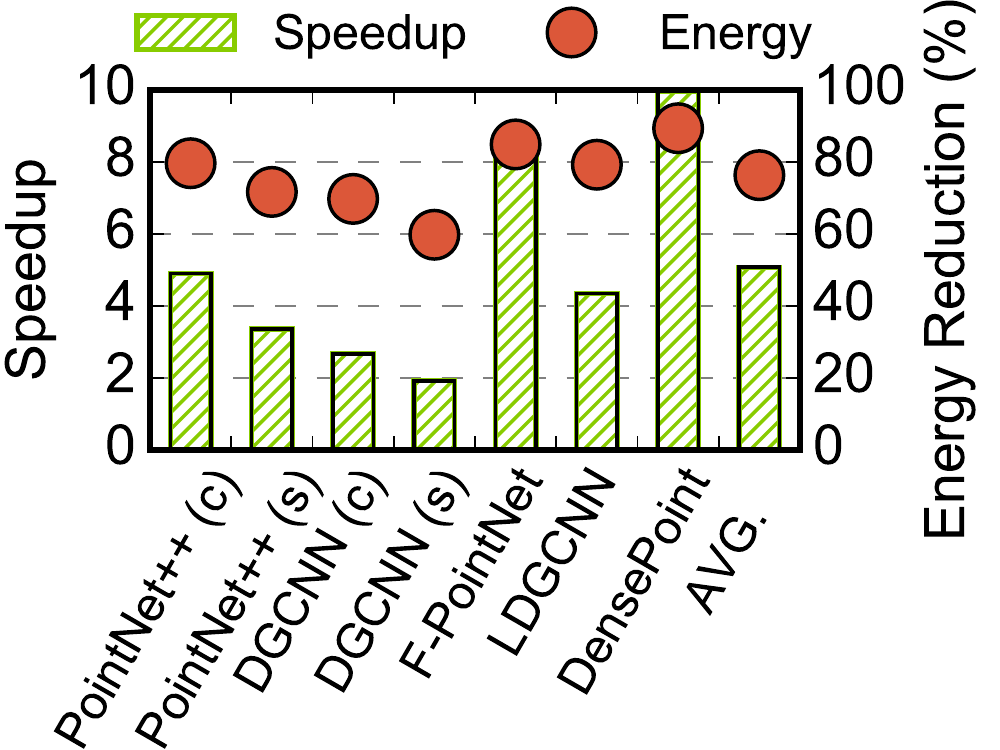}
  \label{fig:systolic_speedup}
  }
  \subfloat[\small{Aggregation.}]
  {
  \includegraphics[width=.48\columnwidth]{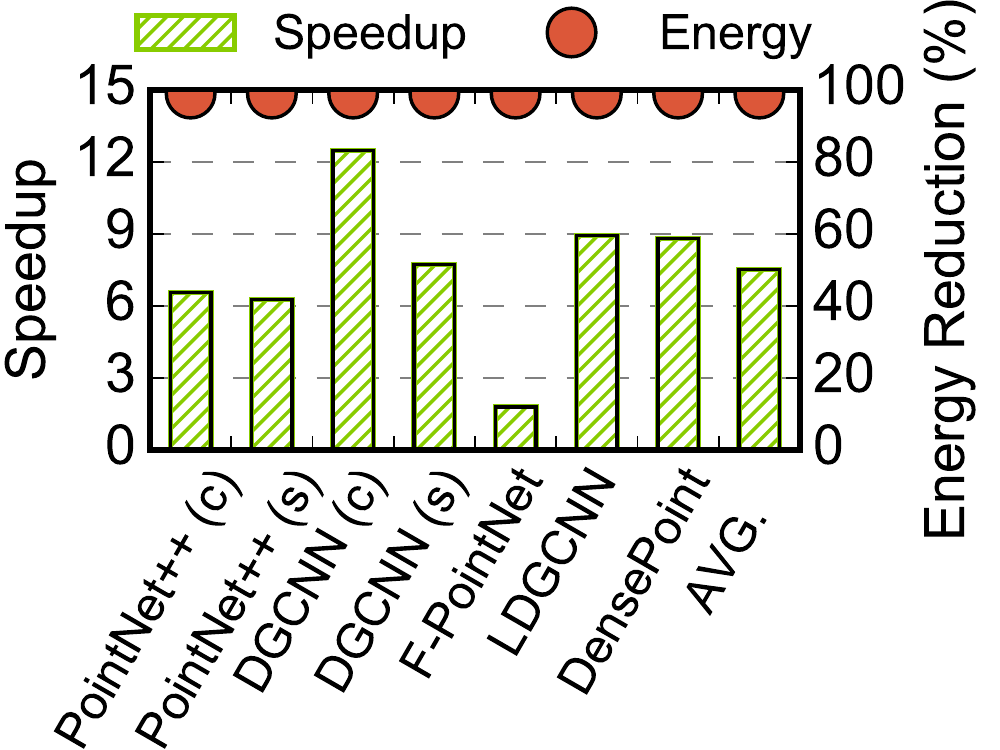}
  \label{fig:group_speedup}
  }
  \caption{Speedup and energy savings on feature computation and aggregation.}
  \label{fig:perf_details}
\end{figure}

\paragraph{Software} The delayed-aggregation algorithm alone without AU support, i.e., \sys{\proj-SW}, has a $1.3\times$ speedup and 22\% energy saving over the baseline. The main contributor of the improvements is optimizing the MLPs in feature computation. \Fig{fig:systolic_speedup} shows the speedups and energy savings of the delayed-aggregation algorithm on feature computation. On average, the feature computation time is reduced by $5.1\times$ and the energy consumption is reduced by 76.3\%.

The large speedup on feature computation does not translate to similar overall speedup, because feature computation time has already been significantly reduced by the NPU, leaving less room for improvement. In fact, our GPU+NPU baseline is about 1.8$\times$ faster (\Fig{fig:hw_speedup}) and consumes 70\% less energy compared to the GPU (\Fig{fig:hw_energy}). The increased workload of aggregation also adds to the overhead, leading to overall lower speedup and energy reduction than on GPU.

\paragraph{Hardware} With the AU hardware, \sys{\proj-HW} boosts the speedup to $1.9 \times$ (up to $3.6\times$) and reduces the energy consumption by 37.6\% (up to $92.9\%$). DGCNN (s) has the least speedup because it has the least aggregation time (\Fig{fig:runtime_dist}), thus benefiting the least from the AU hardware.


\Fig{fig:perf_details} shows the speedup and energy reduction of aggregation over the baseline (which executes aggregation on the GPU). Overall, \sys{\proj-HW} reduces the aggregation time by $7.5\times$ and reduces the energy by 99.4\%. The huge improvements mainly come from using a small memory structure customized to the data access patterns in aggregation.

On average, 27\% (max 29\%) of PFT buffer accesses are to serve previous bank conflicts. The total time spent on PFT buffer accesses is 1.5$\times$ of the ideal case without bank conflicts. Empirically we do not observe pathological cases.

The AU's speedup varies across networks. For instance, the speedup on PointNet++ (c) is over 3$\times$ higher than that of F-PointNet. This is because the speedup decreases as bank conflict increases; bank conflicts occur more often when neighbor search returns more neighbors. The neighbor searches in PointNet++ (c) mostly return 32 neighbors, whereas neighbor searches in F-PointNet return mostly 128 neighbors, significantly increasing the chances of bank conflicts. This also explains why PointNet++ (c) has overall higher speedup than F-PointNet (\Fig{fig:hw_speedup}).

\subsection{Results with Neighbor Search Engine (NSE)}
\label{sec:eval:nse}

From the evaluations above, it is clear that the improvements of \proj will ultimately be limited by the neighbor search overhead, which \proj does not optimize and becomes the ``Amdahl's law bottleneck.''

To assess the full potential of \proj, we evaluate it in a futuristic SoC that incorporates a dedicated neighbor search engine (NSE) that accelerates neighbor searches. We implement a recently published NSE built specifically for accelerating neighbor searches in point cloud algorithms~\cite{xu2019tigris}, and incorporate it into our SoC model. On average, the NSE provides over 60$\times$ speedup over the GPU. Note that the NSE is \textit{not} our contribution. Instead, we evaluate the potential speedup of \proj if an NSE is available.

\begin{figure}
    \centering
    \includegraphics[width=\columnwidth]{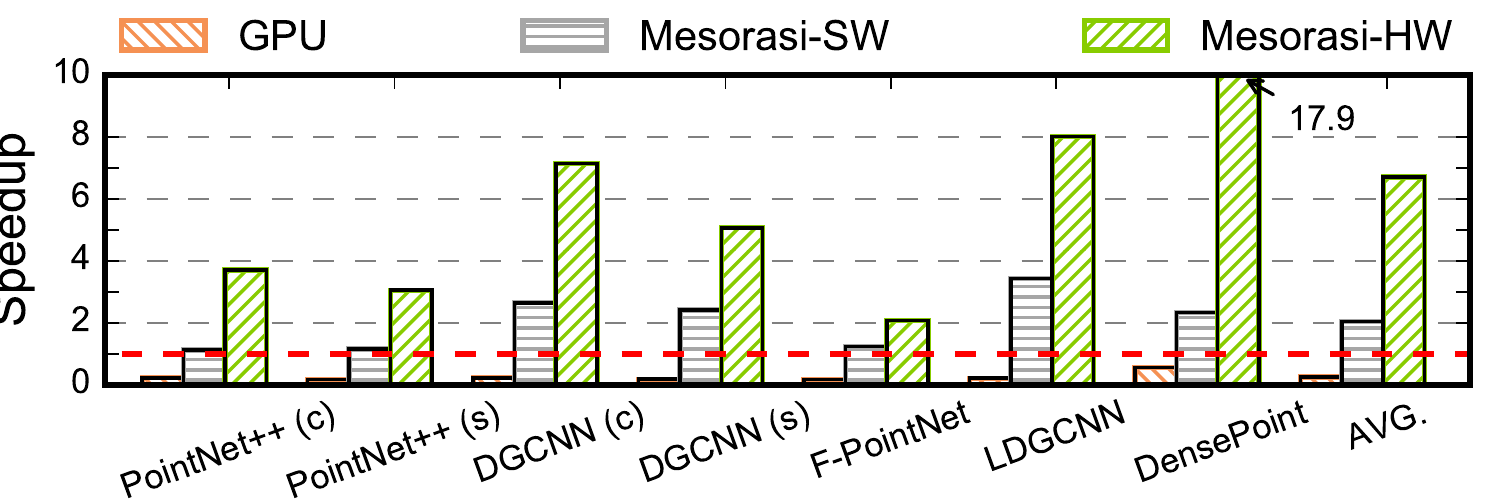}
    \caption{\sys{\proj-SW} and \sys{\proj-HW} speedup over an NSE-enabled SoC (GPU+NPU+NSE), which is 4.0$\times$ faster than the GPU by accelerating both MLP and neighbor search.}
    \label{fig:nse_speedup}
\end{figure}

The speedup of \proj greatly improves when neighbor search is no longer a bottleneck. \Fig{fig:nse_speedup} shows the speedups of \sys{\proj-SW} and \sys{\proj-HW} on the NSE-enabled SoC. On average, \sys{\proj-SW} achieves $2.1\times$ speedup and \sys{\proj-HW} achieves $6.7\times$ speedup. The two DGCNN networks have particularly high speedups because neighbor search contributes heavily to their execution times (\Fig{fig:motivation-dist}).

\subsection{Sensitivity Study}
\label{sec:eval:sen}

The evaluations so far are based on one hardware configuration. We now study how the improvements vary with different hardware resource provisions. In particular, we focus on two types of hardware resources: the baseline NPU and the AU augmentation. Due to the page limit, we show only the results of PointNet++ (s). The general trend holds.

\begin{figure}[t]
\centering
\begin{minipage}[t]{0.48\columnwidth}
  \centering
  \includegraphics[width=\columnwidth]{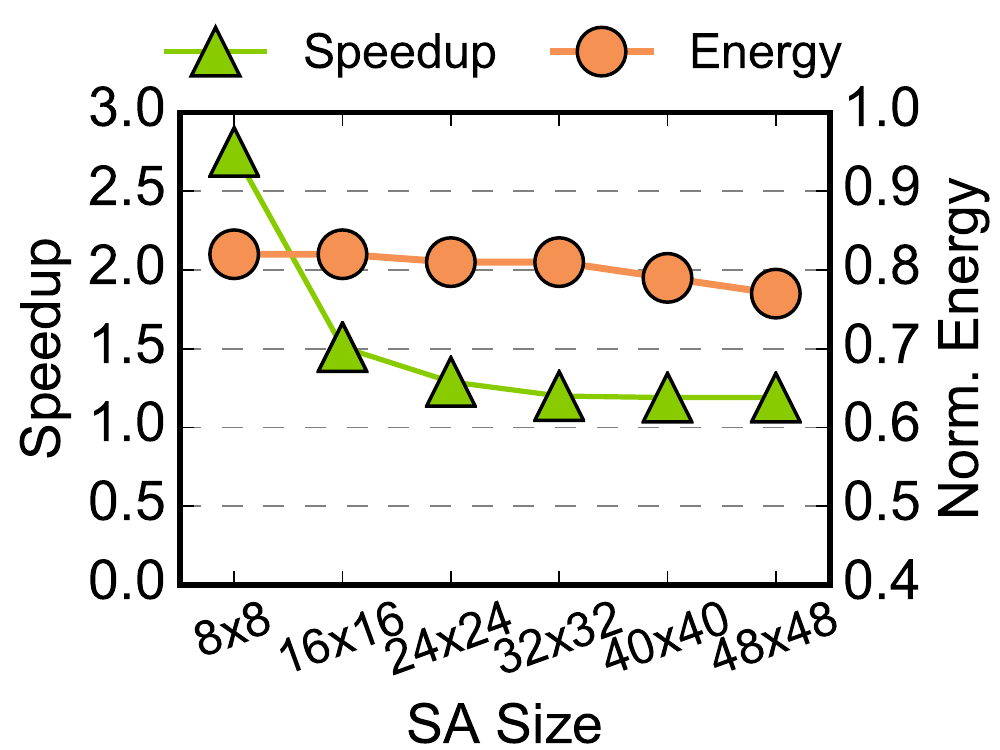}
  \caption{Sensitivity of the speedup and energy to the systolic array size.}
  \label{fig:sa_profile}
\end{minipage}
\hspace{2pt}
\begin{minipage}[t]{0.48\columnwidth}
  \centering
  \includegraphics[width=\columnwidth]{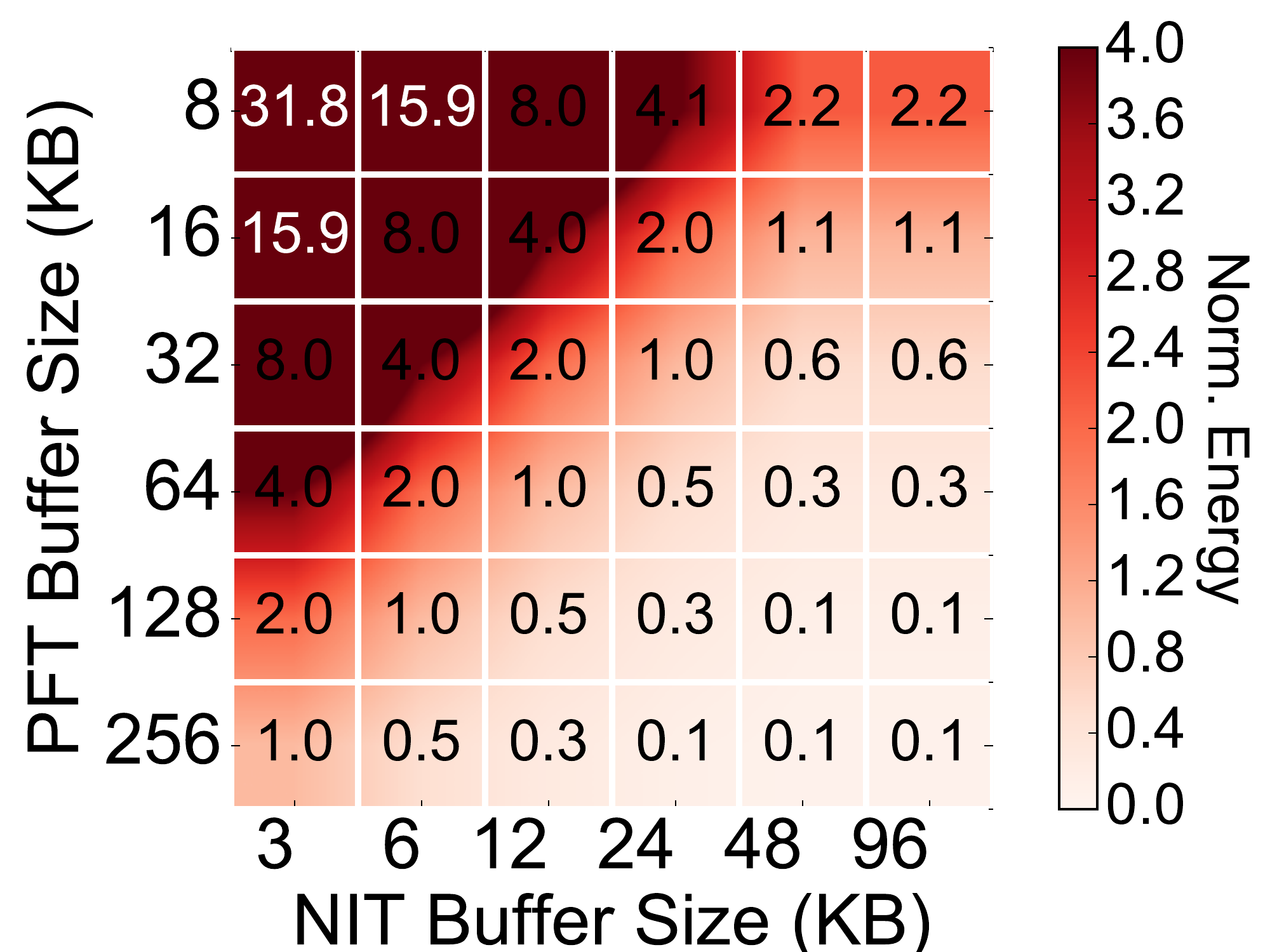}
  \caption{Sensitivity of AU energy consumption to the NIT/PFT buffer sizes.}
  \label{fig:au_profile}
\end{minipage}
\end{figure}

\paragraph{NPU} We find that \proj has higher speedups when the NPU is smaller. \Fig{fig:sa_profile} shows how the speedup and normalized energy of \sys{\proj-HW} over the baseline vary as the systolic array (SA) size increases from $8 \times 8$ to $48 \times 48$. As the SA size increases, the feature extraction time decreases, and thus leaving less room for performance improvement. As a result, the speedup decreases from 2.8$\times$ to 1.2$\times$.

Meanwhile, the energy reduction increases from 17.7\% to $23.4\%$. This is because a large SA is more likely throttled by memory bandwidth, leading to overall higher energy.


\paragraph{AU} We find that the AU energy consumption is sensitive to the NIT and PFT buffer sizes. \Fig{fig:au_profile} shows the AU energy under different NIT and PFT buffer sizes. The results are normalized to the nominal design point described in \Sect{sec:exp} (i.e., a 64 KB of PFT and a 12 KB NIT).

The energy consumption increases as the PFT and NIT buffer sizes decrease. In an extremely small setting with an 8 KB PFT buffer and a 3 KB NIT buffer, the AU energy increases by 32$\times$, which leads to a 5.6\% overall energy increase. This is because a smaller PFT buffer leads to more PFT partitions, which increases NIT buffer energy since each NIT entry must be read once per partition. Meanwhile, a smaller NIT requires more DRAM accesses, whose energy dominates as the PFT buffer becomes very small. On the other extreme, using a 256 KB PFT buffer and a 96 KB NIT buffer reduces the overall energy by 2.0\% while increasing the area overhead by 4$\times$. Our design point balances energy saving and area overhead.

\section{Related Work}
\label{sec:related}

\paragraph{Point Cloud Analytics} Point cloud has only recently received extensive interests. Unlike conventional image/video analytics, point cloud analytics requires considerably different algorithms due to the unique characteristics of point cloud data. Most of the recent work focuses on the accuracy, exploring not only different data representation (e.g., 2D projection~\cite{qi2016volumetric}, voxelization~\cite{wu20153d, riegler2017octnet}, and raw points~\cite{qi2017pointnet++, wang2019dynamic}), but also different ways to extract features from points~\cite{simonovsky2017dynamic, wu2019pointconv, qi2017pointnet++, wang2019dynamic}. Our delayed-aggregation primitive can be thought of as a new, and efficient, way of extracting features from raw points.

\proj focuses on improving the efficiency of point cloud algorithms while retaining the high accuracy. In the same vein, PVCNN~\cite{liu2019point} combines point-based and voxel-based data representations in order to boost compute and memory efficiency. Different but complementary, \proj focuses on point-based neural networks. While PVCNN is demonstrated on GPUs, \proj not only directly benefits commodity GPUs, but also incorporates systematic hardware support that improves DNN accelerators.

Prior work has also extensively studied systems and architectures for accelerating neighbor search on GPU~\cite{Qiu2009GPU, Gieseke2014Buffer}, FPGA~\cite{Winterstein2013FPGA, Kuhara2013An, Heinzle2008A}, and ASIC~\cite{xu2019tigris}. Neighbor search contributes non-trivial execution time to point cloud networks. \proj hides, rather than reduces, the neighbor search latency, and directly benefits from faster neighbor search.

\paragraph{GNNs} Point cloud applications bear some resemblance to GNNs. After all, both deal with spatial/geometric data. In fact, some point cloud applications are implemented using GNNs, e.g., DGCNN~\cite{wang2019dynamic}.

However, existing GNN accelerators, e.g., HyGCN~\cite{yan2020hygcn}, are insufficient in accelerating point cloud applications. Fundamentally, GNN does not require explicit neighbor search (as a vertex's neighbors are explicitly encoded), but neighbor search is a critical bottleneck of all point cloud applications, as points are arbitrarily spread in 3D space. Our design hides the neighbor search latency, which existing GNN accelerators simply do not optimize for. In addition, \proj minimally extends conventional DNN accelerators instead of being a new accelerator design, broadening its applicability in practice.

From \Fig{fig:motivation-dist}, one might notice that $\cA$ in point cloud networks is much faster than $\cF$, which is the opposite in many GNNs~\cite{yan2020hygcn}. This is because $\cF$ in point cloud applications does much more work than $\cA$, opposite to GNNs. In point cloud application, $\cA$ simply gathers neighbor feature vectors, and $\cF$ operates on neighbor feature vectors (MLP on each vector). In contrast, $\cA$ in GNNs gathers and reduces neighbor feature vectors to one vector, and $\cF$ operates on the reduced vector (MLP on one vector).


\paragraph{Domain-Specific Accelerator} Complementary to improving generic DNN accelerators, much of recent work has focused on improving the DNN accelerators for specific application domains such as real-time computer vision~\cite{buckler2018eva2, feng2019asv, zhu2018euphrates}, computational imaging~\cite{huang2019ecnn, mahmoud2018diffy}, and language processing~\cite{riera2018computation}. The NPU in the \proj architecture is a DNN accelerator specialized to point cloud processing. \proj also extends beyond prior visual accelerators that deal with 2D data (images and videos)~\cite{mahmoud2017ideal, leng2019energy, zhang2017race, mazumdar2017exploring, zhang2019distilling, dejaview} to 3D point clouds.

To keep the modularity of existing SoCs, \proj relies on the DRAM for inter-accelerator communication. That said, \proj could benefit from more direct accelerator communication schemes such as VIP~\cite{nachiappan2016vip} and Short-circuiting~\cite{yedlapalli2014short}. For instance, the NIT could be directly communicated to the NIT buffer from the GPU through a dedicated on-chip link, pipelining neighbor search with aggregation.




\section{Conclusion}
\label{sec:conc}

With the explosion of 3D sensing devices (e.g., LiDAR, stereo cameras), point cloud algorithms present exciting opportunities to transform the perception ability of future intelligent machines. \proj takes a systematic step toward efficient point cloud processing. The key to \proj is the delayed-aggregation primitive that decouples neighbor search with feature computation and significantly reduces the overall workload. Hardware support maximizes the effectiveness of delayed-aggregation. The potential gain is even greater in future SoCs where neighbor search is accelerated.



\def\IEEEbibitemsep{0pt plus .5pt}
\bibliographystyle{IEEEtranS}
\interlinepenalty=10000
\bibliography{refs}

\end{document}